\documentclass[preprint,12pt]{elsarticle}
\usepackage{amssymb}
\usepackage{amsmath}
\usepackage{cite} 
\usepackage[colorlinks=true,linkcolor=blue,citecolor=blue,urlcolor=blue]{hyperref}
\usepackage{tabularx}
\usepackage{adjustbox} 
\usepackage{diagbox}
\usepackage{multirow}
\usepackage{booktabs}
\usepackage{multirow}
\usepackage[normalem]{ulem}
\useunder{\uline}{\ul}{}
\usepackage{amsmath}
\usepackage{graphicx}
\usepackage[utf8]{inputenc}
\usepackage{geometry}
\usepackage{enumitem}
\usepackage{lipsum}

\makeatletter
\def\ps@pprintTitle{%
 \let\@oddhead\@empty
 \let\@evenhead\@empty
 \let\@oddfoot\@empty
 \let\@evenfoot\@empty}
\makeatother

\begin{document}
\begin{frontmatter}
\title{Benchmarking Pathology Foundation Models: Adaptation Strategies and Scenarios} 

\author[label1]{Jaeung Lee}
\author[label1]{Jeewoo Lim}
\author[label1]{Keunho Byeon}
\author[label1]{Jin Tae Kwak\corref{cor1}}
\cortext[cor1]{Corresponding author}

\affiliation[label1]{
    organization={School of Electrical Engineering, Korea University}, 
    addressline={}, 
    city={Seoul},
    postcode={02841}, 
    state={},
    country={Republic of Korea}
}

\begin{abstract}
In computational pathology, several foundation models have recently emerged and  demonstrated enhanced learning capability for analyzing pathology images. However, adapting these models to various downstream tasks remains challenging, particularly when faced with datasets from different sources and acquisition conditions, as well as limited data availability.
In this study, we benchmark four pathology-specific foundation models across 14 datasets and two scenarios—consistency assessment and flexibility assessment—addressing diverse adaptation scenarios and downstream tasks. 
In the consistency assessment scenario, involving five fine-tuning methods, we found that the parameter-efficient fine-tuning approach was both efficient and effective for adapting pathology-specific foundation models to diverse datasets within the same downstream task. 
In the flexibility assessment scenario under data-limited environments, utilizing five few-shot learning methods, we observed that the foundation models benefited more from the few-shot learning methods that involve modification during the testing phase only. 
These findings provide insights that could guide the deployment of pathology-specific foundation models in real clinical settings, potentially improving the accuracy and reliability of pathology image analysis. 
The code for this study is available at \textit{https://github.com/QuIIL\\/BenchmarkingPathologyFoundationModels}.
\end{abstract}


\end{frontmatter}

\section{Introduction}
\label{sec1}
Foundation models have recently gained much attention in computational pathology for their superior capability to handle a wide range of downstream tasks \citep{zhang2023challenges}. 
In computational pathology, there are numerous downstream tasks including image classification, segmentation, and registration \citep{zhang2023text, graham2019hover}. Traditionally, we build independent and task-specific models to tackle these tasks, of which each requires sufficient high-quality data for model training and evaluation.
However, this conventional approach poses substantial challenges due to the need for large, labeled datasets and the time-consuming nature of developing and fine-tuning each task-specific model.
With the increase in the number of datasets and advances in model architecture and learning techniques, several pathology-specific foundation models are available \citep{wang2022transformer, kang2023benchmarking, filiot2023scaling, chen2024towards, vorontsov2023virchow, dippel2024rudolfv, lu2024visual}. Despite their success and potential, our understanding of these pathology-specific foundation models is limited. It remains unclear how to best utilize these models for the specific downstream tasks. 
Therefore, there is a demand to set up benchmarks for the pathology-specific foundation models and to gain further insights into these models.
\vspace{12pt}

Developing pathology-specific foundation models generally require four steps: 1) \emph{Model Selection}: select a base model that learns the general knowledge from pathology data; 2) \emph{Data Preparation}: prepare large-scale pathology datasets to teach the model; 3) \emph{Optimization}: identify a learning strategy to teach the model using the data; and 4) \emph{Evaluation}: investigate the trained model on various use cases. 
Specifically, first, one needs to choose the base model among a great deal of artificial intelligence or machine learning models. Among various models, the existing foundation models often adopt Transformers as the base model for the enhanced learning ability especially from a large amount of data. Second, one needs to obtain large-scale datasets to train the chosen model. Many of the existing works utilize pathology images that are publicly available such as those from The Cancer Genome Atlas (TCGA) \citep{tomczak2015review} which contains ~29,000 whole slide images (WSIs) from 25 anatomic sites and covering 32 cancer subtypes with differing pathological conditions and image qualities. Some others adopt their own private pathology image datasets that are used either independently \citep{chen2024towards, vorontsov2023virchow, dippel2024rudolfv} or in combination with TCGA \citep{wang2022transformer, kang2023benchmarking}. Third, one needs to identify a suitable learning strategy to optimize the base model on the large-scale datasets. The existing models are mainly optimized or trained using self-supervised learning (SSL). SSL is a powerful technique that enables learning useful representations/knowledge from the input data with and without data annotations by utilizing the intrinsic property or structure of the data \citep{caron2021emerging, grill2020bootstrap, zhou2021ibot}. This is especially beneficial for pathology images due to the exceptional size of WSIs which are Giga-pixel sized and the scarcity and difficulty of data annotation. Fourth, upon the completion of the optimization, one needs to assess the performance of the foundation models. The existing works evaluate various downstream tasks across different organs and diseases by fully or partially adjusting the weights of the optimized (or pre-trained) foundation models. The number and type of downstream tasks differ one from the other and the performance varies depending on the tasks and the foundation models \citep{chen2024towards, vorontsov2023virchow, dippel2024rudolfv, lu2024visual}.
\vspace{12pt}

Most pathology-specific foundation models are built based upon Transformers trained on public and/or private large-scale datasets using variants of SSL and evaluated on various downstream tasks via fine-tuning. Several research efforts have been made to further improve the efficiency and effectiveness of these foundation models. These, by and large, involve the advancement in architecture of the base model, the inclusion of larger and more diverse datasets, and the enhancement of SSL or other learning algorithms similar to the development trends observed in other artificial intelligence models. Along with these observations, we have noticed that the previous studies have not fully explored the impact of the adaptation strategies. Most existing works adopt linear probing or fine-tuning to adapt the foundation models to downstream tasks. Various adaptation techniques are available, but their effects on the foundation models and downstream tasks remain unknown. Moreover, the downstream tasks usually include single in-domain datasets, which is insufficient for comprehensively investigating the robustness of the foundation models. 
Hence, two critical questions naturally arise: 1) \emph{are the existing models are strong enough to serve as the de facto foundation models?} 2) \emph{what is the optimal strategy to use or adapt the foundation models to downstream tasks?}\vspace{12pt}

To address these two questions and to deepen our understanding of the pathology-specific foundation models, we conduct an in-depth analysis of the foundation models and their behavior under various settings and conditions. Specifically, we employ four pathology-specific foundation models including CTransPath \citep{wang2022transformer}, Lunit \citep{kang2023benchmarking}, Phikon \citep{filiot2023scaling}, and UNI \citep{chen2024towards} and adapt each of the four models to various downstream tasks with 14 datasets and 4 tasks under two experimental scenarios for consistency assessment and flexibility assessment. In the consistency assessment scenario, we evaluate how well the foundation models adapt to different datasets within the same task. This involves various fine-tuning strategies such as linear probing, full fine-tuning, partial fine-tuning, and parameter-efficient fine-tuning (PEFT) to identify the most effective approach to adjust the foundation models for carrying out downstream tasks. In the flexibility assessment scenario, we examine how well the foundation models adapt to the datasets across varying tasks or domains where we explore FSL techniques to further explore the adaptability of the foundation models. Through these two scenarios, we provide a comprehensive understanding of the practical utility of the foundation models in computational pathology and identify the best practices for their deployment in clinical settings.\vspace{12pt}

Our contributions can be summarized as follows:
\begin{itemize}
\item We conduct a comprehensive benchmarking study of the four pathology-specific foundation models, evaluating their performance across 14 datasets from 5 organs.
\item We assess the utility and capability of the foundation models through consistency and flexibility assessment scenarios, providing insights into their robustness and adaptability to various downstream tasks.
\item In the consistency assessment scenario, we investigate the impact of various fine-tuning strategies, including linear probing, full fine-tuning, partial fine-tuning, and PEFT, on the foundation models and their adaptation to differing datasets within the same tasks.
\item In the flexibility assessment scenario, we explore the generalization capabilities of the foundation models across three distinct adaptation scenarios, such as near-domain, middle-domain, and out-domain adaptations, using various FSL methods under data-limited conditions.
\end{itemize}

\section{Related Works}

\subsection{Self-supervised Learning and Pathology Images}
\label{subsec1}
SSL is a learning paradigm that enables learning from unlabeled or partially labeled data by leveraging self-supervised signals that are generated using the intrinsic structure of data. SSL has demonstrated impressive performance in representation learning for various data types including images \citep{he2020momentum, gidaris2018unsupervised, caron2020unsupervised}, videos \citep{pan2021videomoco, tong2022videomae, han2020self}, and text \citep{devlin2018bert, fang2020cert, li2021selfdoc}. SSL has evolved in various ways. A majority of SSL methods are built based upon contrastive learning, which aims to improve the discriminative representation of data. For instance, SimCLR \citep{chen2020simple} and MoCo \citep{he2020momentum} applied data augmentation techniques to maximize the mutual information between different transformations of the same image, bringing similar image pairs closer in the embedding space and pushing dissimilar pairs farther apart. There are non-contrastive SSL approaches. For example, DINO \citep{caron2021emerging} adopted a teacher-student framework using self-distillation, where both models continuously interact to each other to compute embeddings for two augmented views of the same image. BYOL \citep{grill2020bootstrap} introduced a bi-directional learning mechanism for effective feature extraction without labels. Unlike DINO and BYOL, MAE \citep{luo2022self} utilized an encoder-decoder architecture and reconstructed an original image from its duplicate image where a substantial portion has masked out by minimizing the discrepancy between the two outputs obtained from the original image and the reconstructed image.\vspace{12pt}

SSL has been widely adopted for pathology image analysis due to the ability to effectively utilize unlabeled data. Earlier works tend to focus on enhancing the representation power of a model for specific downstream tasks. For instance, IMPaSh \citep{vuong2022impash} proposed a patch shuffling augmentation method, which shuffles the order of patches to address the domain adaptation problem in colon tissue subtyping tasks between different domains. It generated four different image variations and extract and utilized their features to perform contrastive learning. SD-MAE \citep{luo2022self} utilized MAE to process the output of the encoder as a student and the output of the decoder as a teacher, and applied self-distillation between the encoder and decoder for image classification, cell segmentation, and object detection. Self-Path \citep{koohbanani2021self} adopted SSL using both pathology-specific (magnification prediction, magnification puzzles, and Hematoxylin channel prediction) and pathology-agnostic (rotation, flipping, real/fake prediction, and etc.) information. \citep{srinidhi2022self} utilized the SSL method that leverages the pathology images of multiple magnifications as a single sequence and predicts the relative order of magnifications among internal patches using multi-resolution contextual information. HIPT \citep{chen2022scaling} exploited the hierarchical structure of WSI by generating and aggregating image tokens at various resolutions with the student-teacher knowledge distllation method.\vspace{12pt}

In recent years, SSL are increasingly applied to large-scale pathology image datasets that are unlabeled or sparsely labeled \citep{wang2022transformer, kang2023benchmarking, filiot2023scaling, chen2024towards, vorontsov2023virchow, dippel2024rudolfv}. These studies typically leverage publicly available databases such as The Cancer Genome Atlas (TCGA). By training models on extensive unlabeled datasets, these approaches aim to harness the intrinsic patterns and features inherent to pathology images without the need for explicit annotations. Recent developments in the pathology-specific foundation models are generally built based upon SSL and have demonstrated the significant potential to serve as the foundation for various downstream tasks. SSL allows these foundation models to learn robust representations transferable between different pathology tasks, including tumor detection, grading, and subtyping, significantly reducing the need for extensive labeled datasets. For instance, Phikon \citep{filiot2023scaling} employed TCGA to learn pathological feature representations by applying the iBOT \citep{zhou2021ibot} framework, which masks parts of the images and reconstructs the masked portions to learn robust features. CTransPath \citep{wang2022transformer} used TCGA and the Pathology AI Platform (PAIP) \citep{kim2021paip} dataset and proposed an enhanced contrastive learning method that selects positive instances with similar visual information from a memory bank. 
Moreover, several other studies utilize large-scale private datasets. For example, Lunit \citep{kang2023benchmarking} adopted two datasets such as TCGA and TULIP, a private dataset with 1.3 million image patches and utilized data from various fields of view at 20× and 40× magnifications. It employed augmentation techniques specific to pathology images, including random vertical flip and RandStainNA for stain augmentation and DINO \citep{caron2021emerging} for training. 
UNI \citep{chen2024towards} introduced Mass-100k with over 100 million images, trained the ViT-L model, and conducted 34 downstream tasks.
Virchow \citep{vorontsov2023virchow} collected 1.5 million WSIs to train the ViT-H/14 that is evaluated on 26 downstream tasks. The performance of the model was evaluated on 31 different datasets.
RudolfV \citep{dippel2024rudolfv} integrated data from over 15 laboratories, including 134,000 WSIs from 34,000 cases and assesses 12 downstram tasks by using ViT-L model. 
For model training, UNI, Virchow, and RudolfV employed DIONv2 [reference], which involves masked image modeling and self-distillation to learn meaningful representations.

\subsection{Pre-training and Fine-tuning}
\label{subsec2}
As the foundation models, trained either by SSL or other learning paradigms, are applied to downstream tasks, they usually undergo fine-tuning to adjust themselves to specific tasks or problems. Various fine-tuning techniques are available such as linear probing \citep{kornblith2019better}, full fine-tuning, partial fine-tuning, and PEFT. Linear probing is the simplest yet efficient adaptation method, wherein the final linear classifier is trained on the downstream tasks while the rest of the models remain frozen. The simplicity of the method allows for quick adaptation but may not fully utilize the potential of the foundation models. In contrast, full fine-tuning updates the weights across all layers of the models. This approach can offer improved performance for downstream tasks but is computationally expensive. Partial fine-tuning involves adjusting certain layers while keeping others frozen. PEFT is an emerging adaptation method, originated from natural language processing, that can adjust the model in a parameter-efficient fashion. Low-rank adaptation (LoRA) \citep{hu2021lora}, for example, hypothesized that the learnable weights of a model reside on a low intrinsic dimension and proposed to update the rank decomposition of the weights during adaptation. LoRA and its variants have been successfully applied to various problems such as generating radiology texts \citep{liu2023radiology}, retinal Optical Coherence Tomography (OCT) segmentation \citep{fazekas2023adapting}, and multi-organ medical image segmentation \citep{liu2024moe}. Radiology-Llama2 \citep{liu2023radiology} used instruction tuning and LoRA for radiology reports, generating coherent and clinically useful texts, outperforming conventional large language models (LLMs). SAMedOCT \citep{fazekas2023adapting} adapted the Segment Anything Model (SAM) for retinal OCT scans with LoRA for efficient fine-tuning. MOELoRA \citep{liu2024moe} combined Mixture-of-Expert (MOE) models and LoRA for multi-task medical applications, addressing data imbalance and reducing computational costs.

Most foundation models have adopted linear probing for the adaptation to downstream tasks. To the best of our knowledge, PEFT has not been utilized for the adaptation of the pathology-specific foundation models.

\subsection{Few-shot Learning}
\label{subsec3}
FSL is a technique to learn from a limited number of labeled examples, often only a few samples per class. The term “few-shot” emphasizes the efficiency of a model to adapt to new tasks while minimizing exposure to new data. FSL is particularly useful in scenarios where the acquisition of extensive annotated datasets is challenging and/or expensive. The key of FSL is the N-way K-shot learning setting, where a model is trained using only K labeled examples for each of N different classes. It usually employs a training strategy called episodic training, which simulates real learning tasks with limited data. In each episode, the model is exposed to a randomly selected subset of classes known as the support set, each represented by K examples. Then, the model predicts the class label of new examples, designated as a query set.\vspace{12pt}

FSL falls under the learning paradigm of meta-learning, which learns how to learn and adapt to new tasks using a few examples only. Matching Networks (MatchingNet) \citep{vinyals2016matching} divided data in a support set and a query set and trained a model to maximize the similarity between the data in the query set and the corresponding data in the support set. Consequently, during testing, the model was able to dynamically adapt to new data (or classes) by assigning the label of the data from the support set that shows the highest similarity to each test example in the query set. Prototypical Networks (ProtoNet) \citep{snell2017prototypical} computed the central vector (prototype) for each class based on the average of all example embeddings in the support set. To classify a new example from the query set, the model measured the Euclidean distance between the embedding of the query examples and the prototype of each class. The query examples were classified into the pertinent class with the closest prototype. Model-Agnostic Meta-Learning (MAML) \citep{finn2017model} proposed a model to adapt quickly to new tasks through a two-step training process. In the first step, the model was trained to determine a good starting parameter set that can be quickly fine-tuned with a small number of gradient updates for various tasks. In the second step, the trained model was evaluated on new tasks, meanwhile updating the initial parameters to optimize its performance across tasks.\vspace{12pt}

Several research efforts have adopted FSL for pathology image analysis. For example, \citep{shaikh2022artifact} utilized ProtoNet to identify artifacts in pathology images. The extension of ProtoNet via k-means has been proposed to address a data distribution shift due to different scanners \citep{deuschel2021multi}. Furthermore, \citep{shakeri2022fhist} investigated the impact of different FSL methods in clinical environments that deal with various cancer types and histopathological conditions. However, to the best of our knowledge, the effect of FSL on the pathology-specific foundation models has not been investigated. In this study, we adopt various FSL approaches to assess the generalization capability of the pathology-specific foundation models based upon a few examples across various tasks and datasets.\vspace{12pt}

\begin{figure}
\centering
\includegraphics[width=\textwidth,keepaspectratio]{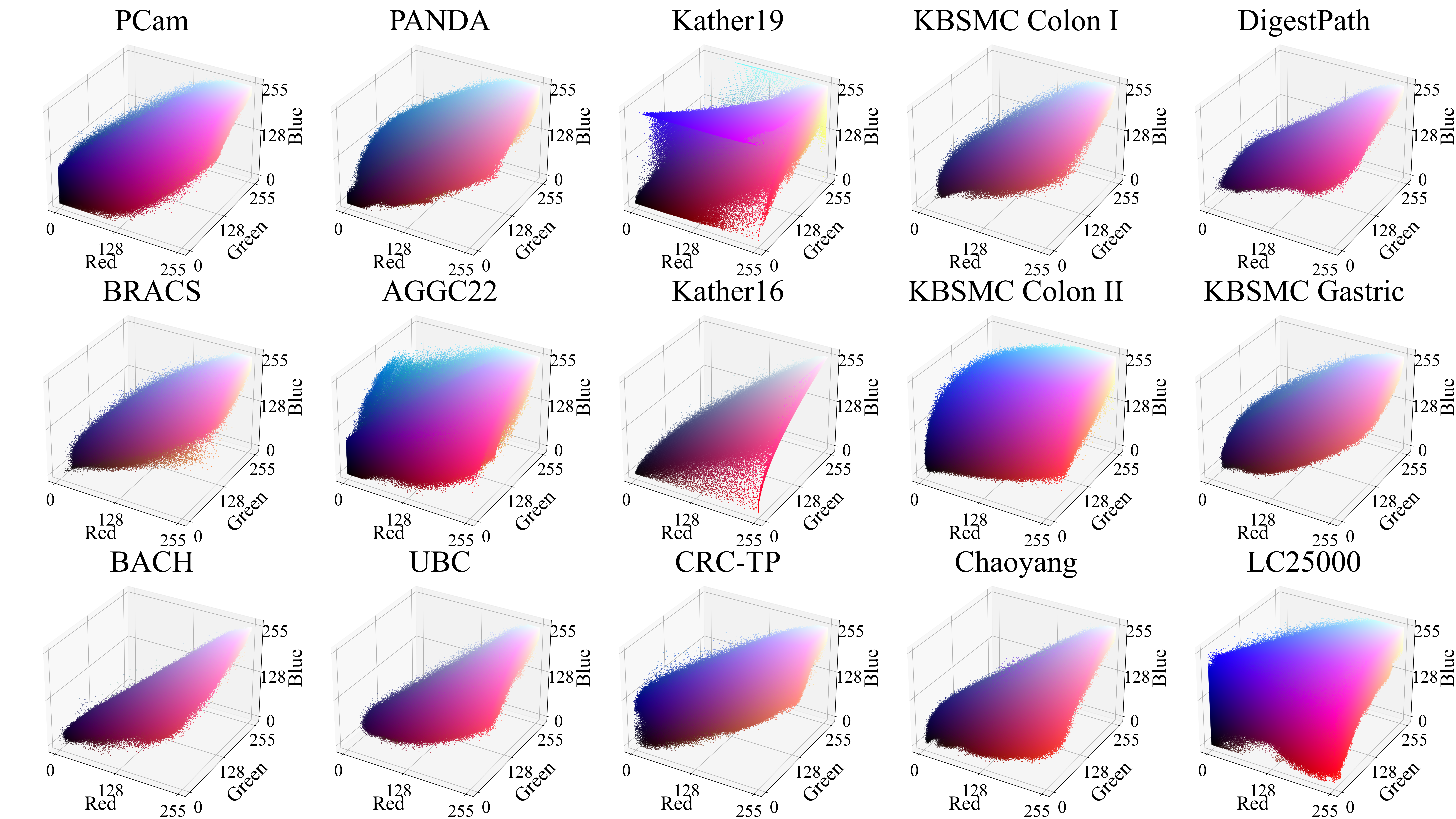}
\caption{3D RGB color distribution of various pathology datasets.}
\label{fig:color_distribution}
\end{figure}

\section{Methodology}
\subsection{Dataset}
\label{subsec4}
We employ 14 publicly accessible pathology datasets to evaluate the performance of the pathology-specific foundation models on consistency and flexibility assessment scenarios. The datasets include pathology images with differing pathological conditions that are acquired from various organs and institutions. Due to differences in acquisition environments and digital scanners, there is substantial variability in data, such as color distribution, which can lead to domain shifts (Fig. \ref{fig:color_distribution}). Table \ref{tab:datasets} demonstrates a summary of the entire datasets.\vspace{12pt}

\textbf{Kather19} \citep{kather2019predicting} comprises 100,000 image patches derived from 86 colorectal WSIs. These image patches have a spatial size of 224×224 pixels and are scanned at a pixel resolution of approximately 0.5µm under 20× magnification. This dataset includes nine distinct colorectal tissue categories: tumor tissue, simple stroma, complex stroma, immune cells, debris, normal colon mucosa, adipose tissue, mucus, and smooth muscle.\vspace{12pt}

\textbf{Kather16} \citep{kather2016multi} contains 5,000 image patches, sourced from anonymized colorectal tissue slides in the pathology archive at the University Medical Center Mannheim, Heidelberg University, Germany. Each image, of size 150×150 pixels, represents one of eight different tissue types: tumor epithelium, simple stroma, complex stroma, immune cells, debris, mucosal glands, adipose tissue, and background.\vspace{12pt}

\textbf{CRC-TP} \citep{javed2020cellular} consists of 196,000 image patches, each of size 150×150 pixels, derived from 20 WSIs of colorectal tissues and scanned at 20× magnification from University Hospitals Coventry and Warwickshire. The dataset categorizes image patches into seven classes: tumor, inflammatory, stroma, complex stroma, necrotic, benign, and smooth muscle.\vspace{12pt}

\begin{table}[t]
\centering
\caption{Overview of pathology datasets with detailed characteristics.}
\label{tab:datasets}
\begin{adjustbox}{max width=\textwidth}
\begin{tabular}{@{}llccccccc@{}}
\toprule
Organ              & Dataset        & \# Classes & \# Samples & \# Patients & \# Patches & Patch size    & Pixel size & FoV \\ \midrule
Breast             & PCam           & 2          & 400        & -           & 327,680    & 96x96         & 0.970µm    & 10×          \\
                   & BACH           & 4          & -          & -           & 14,271     & 512x512       & 0.420µm    & 20×          \\
                   & BRACS          & 7          & 387        & 380         & 4,539      & Variable size & 0.250µm    & 40×          \\ \midrule
Colorectal         & Kather19       & 9          & 86         & 86          & 100,000    & 224x224       & 0.500µm    & 20×          \\
                   & Kather16       & 8          & 10         & -           & 5,000      & 224×224       & 0.495µm    & 20×          \\
                   & CRC-TP         & 7          & 20         & 20          & 196,000    & 150x150       & -          & 20×          \\
                   & KBSMC Colon I   & 4          & 343        & 343         & 9,857      & 1,024x1,024   & 0.247µm    & 40×          \\
                   & KBSMC Colon II & 4          & 45         & 45          & 110,170    & 1,144×1,144   & 0.225µm    & 40×          \\
                   & DigestPath     & 2          & 324        & -           & 107,982    & 512×512       & -          & 20×          \\
                   & Chaoyang       & 4          & -          & -           & 6,160      & 512×512       & -          & 20×          \\ \midrule
Prostate           & PANDA          & 4          & 5,158      & -           & 735,593    & 512×512       & 0.240µm    & 20×          \\
                   & AGGC22         & 5          & 286        & -           & 323,697    & 512×512       & 0.500µm    & 20×          \\
                   & UBC            & 4          & 244        & -           & 17,066     & 690×690       & 0.250µm    & 40×          \\ \midrule
Gastric            & KBSMC Gastric  & 8          & 98         & 98          & 206,136    & 1,024×1,024   & 0.264µm    & 40×          \\ \midrule
Lung \& Colorectal & LC25000        & 5          & -          & -           & 25,000     & 768×768       & -          & -            \\ \bottomrule
\end{tabular}
\end{adjustbox}
\end{table}

\textbf{KBSMC Colon} \citep{le2021joint} comprises 120,123 image patches derived from colorectal WSIs and tissue microarrays (TMAs) from 45 patients. Each patch has a spatial size of 512×512 pixels with a pixel size of 0.2465µm, which is digitized at 40× magnification. The patches are categorized into four classes based on the differentiation level of cancer: benign, well-differentiated, moderately differentiated, and poorly differentiated. The dataset is divided into two parts: KBSMC Colon I and KBSMC Colon II, based on the time of acquisition and choice of digital scanners, introducing variations between the two datasets. KBSMC Colon I is digitized with an Aperio digital slide scanner (Leica Biosystems), while KBSMC Colon II is digitized with a NanoZoomer digital slide scanner (Hamamatsu Photonics K.K.).\vspace{12pt}

\textbf{Chaoyang} \citep{zhu2021hard} consists of 6,160 image patches, of size 512×512 pixels, derived from colon tissues, collected from 324 patients at Chaoyang Hospital, affiliated with Capital Medical University in Beijing. These image patches are scanned at 20× magnification and classified into four categories: normal, serrated, adenocarcinoma, and adenoma.\vspace{12pt}

\textbf{DigestPath} \citep{da2022digestpath} includes a colonoscopy tissue dataset for the automatic segmentation and classification of colorectal tissues. This dataset contains malignant and benign samples. The training set of malignant samples comprises 250 images with pixel-level annotations from 93 WSIs, while the benign samples consist of 410 images from 231 WSIs. All WSIs are scanned at 20× magnification using a KFBIO FK-Pro-120 slide scanner (KFBio).\vspace{12pt}

\textbf{PANDA} \citep{bulten2022artificial} is a dataset for prostate cancer diagnosis and Gleason grading. The dataset consists of 88,199 image patches derived from 5,158 WSIs, digitized at a magnification of 20×. Each patch is 512×512 pixels in size and includes pixel-level annotations that classify the tissue into four categories: benign, grade 3, grade 4, and grade 5.\vspace{12pt}

\textbf{AGGC22} \citep{huo2024comprehensive} is obtained from the training set of the Automated Gleason Grading Challenge 2022. This dataset includes three distinct subsets, digitized at 20× magnification. Two subsets are scanned using an Akoya Biosciences scanner, while the third subset is obtained using six different scanners: Akoya Biosciences, KFBio, Leica, Olympus, Philips, and Zeiss. It comprises 323,697 prostate image patches of size 512 x 512 pixels extracted from 249 whole mount images and 37 biopsy images. The patches are categorized into five classes: stroma, benign, grade 3, grade 4, and grade 5.\vspace{12pt}

\textbf{UBC} \citep{nir2018automatic} is part of the training set of the Gleason2019 challenge. The dataset comprises 17,066 prostate image patches, each contains 690×690 pixels, derived from 244 prostate tissue cores. These samples are digitized with an Aperio digital slide scanner (Leica Biosystems) at a 40× magnification and annotated by six pathologists at the Vancouver Prostate Centre. The patches are categorized into 4 classes: benign, grade 3, grade 4, and grade 5.\vspace{12pt}

\textbf{PCam} \citep{veeling2018rotation} includes 327,680 breast image patches that are extracted from the CAMELYON16 Challenge dataset \citep{bejnordi2017diagnostic}, which contains 400 WSIs of sentinel lymph node sections. These slides are acquired and digitized at two different centers using two different scanners: the Pannoramic 250 Flash II (3DHISTECH Ltd.) with 20× magnification and the NanoZoomer-XR Digital slide scanner C12000-01 (Hamamatsu Photonics K.K) with 40× magnification. The images are undersampled to 10× magnification. Each patch measures 96×96 pixels. The dataset consists of two classes: tumor and normal, with an equal number of patches in each category.\vspace{12pt}

\textbf{BRACS} \citep{brancati2022bracs} includes 547 breast WSIs collected from 189 patients, along with 4,539 regions of interests (ROIs). All slides are scanned with an Aperio AT2 (Leica Biosystems) at 40× magnification. Each ROI is annotated by consensus among three pathologists. The dataset encompasses a range of lesion types, including benign, malignant, and atypical, which are subdivided into seven classes: normal, pathological benign, usual ductal hyperplasia, flat epithelial atypia, atypical ductal hyperplasia, ductal carcinobma in situ, and invasive carcinoma.\vspace{12pt}

\textbf{BACH} \citep{aresta2019bach} consists of 400 pathology image patches of breast tissue. Each patch has 2048×1536 pixels, digitized  at 20× magnification with a Leica SCN400 digital slide scanner (Leica Biosystems). These are collected for the Grand Challenge on Breast Cancer Histology held during the ICIAR 2018 conference. The dataset is divided into four classes: normal, benign, in situ carcinoma, and invasive carcinoma.\vspace{12pt}

\textbf{KBSMC Gastric} \citep{lee2023camel} is derived from 98 gastric WSIs collected from 98 patients, resulting in 206,136 image patches. Each patch has a spatial size 1,024×1,024 pixels, scanned at 40× magnification using an Aperio digital slide scanner (Leica Biosystems) with a pixel size of 0.2635µm. The dataset has eight categories: benign, tubular well-differentiated adenocarcinoma, tubular moderately-differentiated adenocarcinoma, tubular poorly-differentiated adenocarcinoma, gastric carcinoma with lymphoid stroma, papillary carcinoma, mucinous carcinoma, and poorly cohesive carcinoma (including signet ring cell carcinoma and other poorly cohesive types).\vspace{12pt}

\textbf{LC25000} \citep{borkowski2019lung} includes 25,000 image patches categorized into five classes with 5,000 images each, representing colon adenocarcinoma, benign colonic tissue, lung adenocarcinoma, lung squamous cell carcinoma, and benign lung tissue. All images are resized to 768×768 pixels.

\subsection{Foundation Models}
\label{subsec5}
We employ four pathology-specific foundation models including CTransPath, Lunit, Phikon, and UNI that are pre-trained on a large collection of pathology image datasets using SSL-based methods.\vspace{12pt}

\textbf{CTransPath} \citep{wang2022transformer} adopts a hybrid architecture combining a convolutional neural network (CNN) and a multi-scale Swin Transformer facilitating a collaborative local-global feature extraction. CTransPath is pre-trained on a large unlabeled dataset of pathology images from TCGA and PAIP, comprising approximately 15 million image patches cropped from over 30,000 WSIs by leveraging the semantically relevant contrastive learning (SRCL).\vspace{12pt}

\textbf{Lunit} \citep{kang2023benchmarking} utilizes DINO \citep{caron2021emerging} to train ViT-S on 32.6 million image patches obtained from TCGA and TULIP. TULIP is a private dataset that contains 13.6 million image patches of size 512×512 pixels. It adopts domain-specific data augmentation and field of view adjustments to optimize the model for high accuracy and reliability.\vspace{12pt}

\textbf{Phikon} \citep{filiot2023scaling} adopts the iBOT \citep{zhou2021ibot} framework, which is a SSL approach that uses Masked Image Modeling (MIM), to train ViT-B on a dataset comprising over 40 million pathology images across 16 different cancer types originated from TCGA. Given an image, MIM randomly masks some regions and reconstructs the masked regions to learn useful and meaningful representations.\vspace{12pt}

\textbf{UNI} \citep{chen2024towards} employs DINOv2, which is a SSL approach that uses self-distillation and MIM, to train ViT-L n a private large-scale dataset named Mass-100K. Self-distillation aligns the prediction distributions between the student and teacher networks to enhance learning stability, while MIM focuses on reconstructing masked regions of an image to capture meaningful representations.  The Mass-100K dataset comprises over 100 million images from more than 100,000 WSIs across 20 major tissue types.\vspace{12pt}

\begin{table}[t]
\centering
\caption{Summary of four pathology-specific foundation models.}
\label{tab:models}
\begin{adjustbox}{max width=\textwidth}
\begin{tabular}{@{}lccccccc@{}}
\toprule
Model      & Backbone   & Training method & Dataset    & \# WSIs (K) & \# Patches (M) & Patch size  & FoV      \\ \midrule
CTransPath & CNN + Swin & SRCL            & TCGA+PAIP  & 32          & 15.4           & 1,024×1,024 & 20×      \\
Lunit      & ViT-S      & DINO            & TCGA+TULIP & 36          & 32.6           & 512×512     & 20×, 40× \\
Phikon     & ViT-B      & iBOT            & TCGA       & 6           & 43.3           & 224×224     & 20×      \\
UNI        & ViT-L      & DINOv2          & MASS-100k  & 100         & 100            & 256×256     & 20×      \\ \bottomrule
\end{tabular}
\end{adjustbox}
\end{table}

\begin{figure}[ht]
\centering
\includegraphics[width=\textwidth,keepaspectratio]{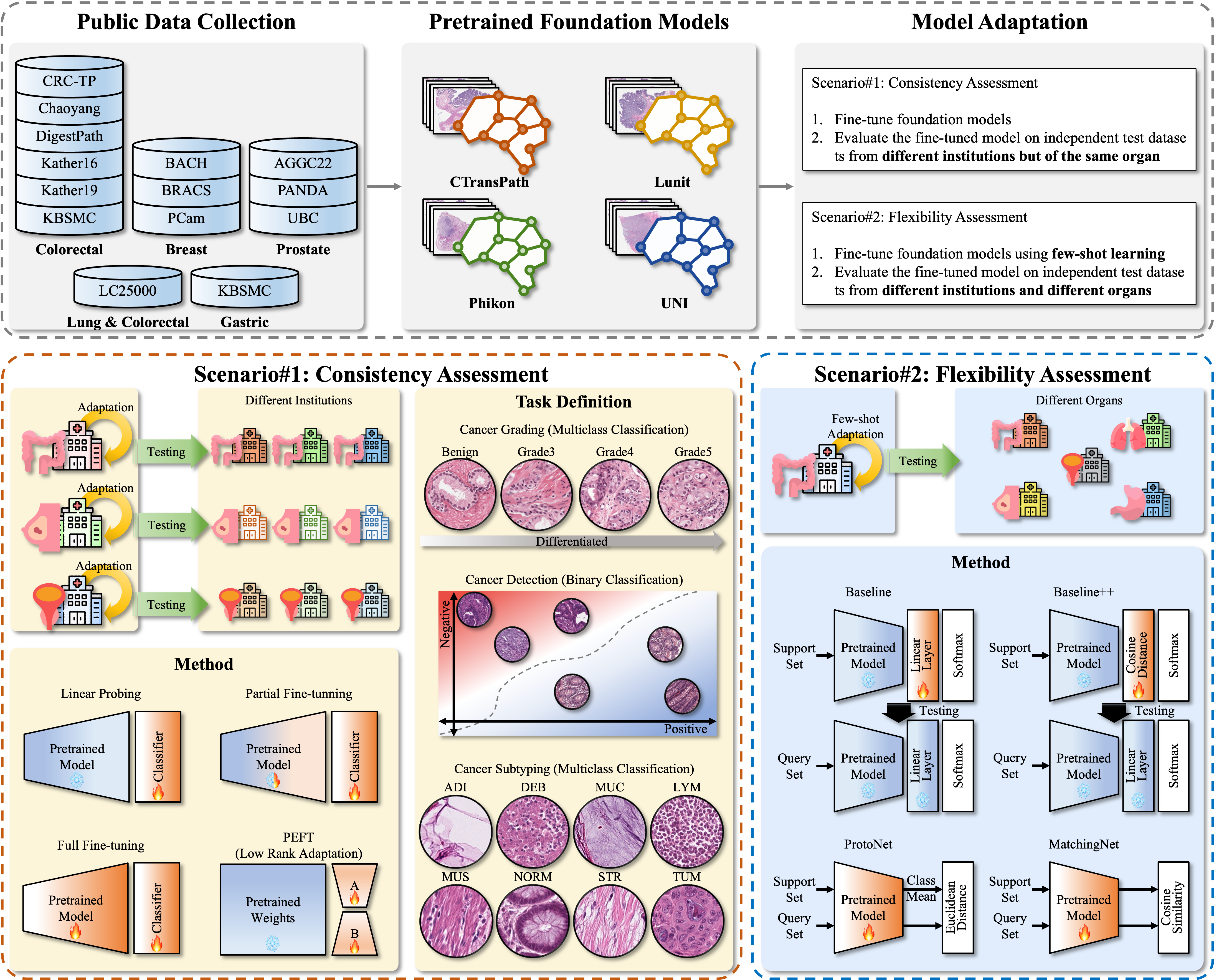}
\caption{Overview of our benchmark study.}
\label{fig:benchmark_overview}
\end{figure}

\subsection{Benchmark Study Design: Consistency Assessment Scenario}
\label{subsec6}
The consistency assessment scenario is designed to investigate various fine-tuning methods on the pathology-specific foundation models and to identify the most suitable fine-tuning method for seamless and consistent adaptation to downstream tasks. Given a downstream task, we utilize multiple datasets acquired from various institutions and acquisition environments to assess the quality of fine-tuning methods in the context of generalization ability and domain shift issues.

\subsubsection{Task Definition}
\label{subsubsec1}
We define four tasks with multiple datasets from three organs — breast, prostate, and colon.\vspace{12pt}

\textbf{Breast Cancer Detection.} Breast cancer detection involves the binary classification of breast tissues into tumor and non-tumor categories. Three breast cancer datasets are considered: PCam, BRACS, and BACH. PCam is divided into a training subset, a validation subset, and a test subset. We use the training subset to fine-tune the foundation models and the validation subset to choose the best checkpoint of the models that is applied to the test subset and the other two datasets (BRACS and BACH). BRACS and BACH are re-grouped into two categories, i.e., tumor and non-tumor, and are designated as independent unseen datasets to assess the generalization performance of the foundation models in breast cancer detection. In BRACS, each ROI has a width of $\leq$ 17,611 pixels and a height of $\leq$ 13,462 pixels. For our experiments, we extract patches of 256×256 pixels from these ROIs.\vspace{12pt}

\textbf{Colorectal Cancer Detection.} Colorectal cancer detection is a binary classification task that distinguishes tumor colorectal tissues from and non-tumor colorectal tissues. In this task, we employ four colorectal cancer datasets: KBSMC Colon I, KBSMC Colon II, DigestPath, and Chaoyang. Following \citep{le2021joint}, we split KBSMC Colon I into a training subset, a validation subset, and a test subset for model fine-tuning and evaluation as used above. The fine-tuned foundation models are evaluated on the test subset of KBSMC Colon I and the other three datasets (KBSMC Colon II, DigestPath, and Chaoyang). We note that KBSMC Colon I and KBSMC Colon II were acquired from the same institution but were digitized at different times using different digital slide scanners.\vspace{12pt}

\textbf{Colorectal Sub-Typing.} Colorectal sub-typing is a task to classify colorectal tissues into seven sub-types including tumor, inflammatory, stroma, complex stroma, necrotic, benign, and smooth muscle. We employ Kather19 for fine-tuning and use Kather16 and CRC-TP as independent unseen datasets. Kather19 was split into 70\% training set, 15\% validation set, and 15\% testing set. Since Kather19 and Kather16 have nine and eight categories respectively, we follow \citep{abbet2021self} to re-group both datasets into seven categories such as adipose, background, debris, lymphocytes, normal, stroma, and tumor. We accomplish this by grouping stroma/muscle and debris/mucus as stroma and debris, respectively. Similarly, stroma/muscle in the CRC-TP dataset are grouped as stroma. Due to inconsistencies in the definition of complex stroma between Kather16 and CRC-TP, complex stroma is excluded from both datasets.\vspace{12pt} 

\textbf{Prostate Cancer Grading.} Prostate cancer grading is a task that aims to classify prostate cancer tissues into benign, grade 3, grade 4, and grade 5. For this task, we employ PANDA for fine-tuning, which is split into a training subset, a validation subset, and a test subset, following \citep{vuong2023moma}. Then, we use UBC and AGGC22 for independent evaluation.\vspace{12pt}

\subsubsection{Fine-tuning Methods}
\label{subsubsec2}
We consider four fine-tuning methods, such as fully supervised learning, linear probing, full fine-tuning, partial fine-tuning, and PEFT and one training method, fully supervised learning, to adjust the pathology-specific foundation models for each downstream task. Fully supervised learning involves initializing all weights of the foundation models and training from scratch. This method does not utilize the pre-trained weights and is used to evaluate the \texttt{Baseline} performance of the models without any pre-trained knowledge. Linear probing freezes the entire layers in the foundation models except the last classification layer, which is tailored for downstream tasks. Full fine-tuning updates the weights across all layers of the models. Partial fine-tuning adjusts the weights in the bottom 50\% of the layers in the models while freezing the top 50\% of the layers. For PEFT, we adopt LoRA \citep{hu2021lora}, which introduces a low-rank decomposition of the weight matrices for parameter-efficient fine-tuning. Suppose that we are given a weight matrix $W_0 \in \mathbb{R}^{d \times k}$. We update $W_0$ with a low-rank decomposition as follows: $W = W_0 + \Delta W = W_0 + BA$ where $B \in \mathbb{R}^{d \times r}$, $A \in \mathbb{R}^{d \times k}$, and the rank $r \ll \min(d, k)$. LoRA is adopted to adjust each self-attention layer of the foundation models. $r$ is set to 8.

\subsubsection{Implementation Details}
\label{subsubsec3}
We train and fine-tune all models using the Adam optimizer with default parameter values (\( \beta_1 = 0.9 \), \( \beta_2 = 0.999 \), and \( \epsilon = 1.0e-8 \)) and 256 batch size. All input images are resized to 224x224 pixels. Cross-entropy loss is adopted for supervised learning during fine-tuning. We apply color jitter, random horizontal flipping, and random resizing cropping as data augmentation techniques. All the models are implemented on the PyTorch platform and executed on a workstation with four RTX A6000 GPUs.

\subsection{Benchmark Study Design: Flexibility Assessment Scenario}
\label{subsec7}
The flexibility assessment scenario evaluates generalization capability of the pathology-specific foundation models to adapt to the downstream tasks with a limited number of examples for fine-tuning, in which the traditional fine-tuning methods are not applicable. Instead of the four fine-tuning methods in the consistency assessment scenario, we explore a number of FSL methods to adjust and apply the pathology-specific foundation models to new tasks involving different organ types. Specifically, the foundation models are adjusted on a meta training dataset $X_{Meta-train}$, evaluated on a meta validation data $X_{Meta-val}$ to select the best checkpoints, and then applied them to a meta test dataset $X_{Meta-test}$.
$X_{Meta-train}$, $X_{Meta-val}$, and $X_{Meta-test}$ are collected from different organs and/or institutes with no overlap in their corresponding label spaces: $Y_{Meta-train}$ , $Y_{Meta-val}$, and $Y_{Meta-test}$ ($Y_{Meta-train} \cap Y_{Meta-val} \cap Y_{Meta-test} = \emptyset$). This indicates that the foundation models are asked to handle a new task with unseen classes. 

For FSL-based adaptation, we utilize an $N$-way $K$-shot framework. This framework constructs a series of episodes by randomly selecting $N$ classes from the label spaces ($Y_{Meta-train}$, $Y_{Meta-val}$, or $Y_{Meta-test}$) and $K$ examples from each class to produced a support set. Each episode also contains a query set, which consists of $Q$ samples from the same classes as those in the support set. The support set and query set are individually and independently formed for each of the meta datasets ($X_{Meta-train}$, $X_{Meta-val}$, or $X_{Meta-test}$).
We use CRC-TP and Kather19 as $X_{Meta-train}$ and $X_{Meta-val}$, respectively. Though both datasets consist of colorectal tissue samples, they were acquired from different institutions with differing tissue types. For  $X_{Meta-test}$, we employ five datasets, including KBSMC Colon, LC25000, PANDA, KBSMC Gastric, and BACH, acquired from five distinct organs such as colon, lung, prostate, gastric, and breast.
These datasets serve as the basis for three adaptation tasks, each of which introduces various sources of variability such as differences in organ type, labeling criteria, and institutions where the datasets were collected. These variations are crucial for assessing the robustness and generalization capabilities of foundation models across diverse datasets and tasks.


\subsubsection{Task Definition}
\label{subsubsec4}

Following FHist \citep{deuschel2021multi}, we explore the flexibility of the pathology-specific foundation models by assessing three distinct adaptation tasks: 1) near-domain adaptation, 2) middle-domain adaptation, and 3) out-domain adaptation. 
The details of the adaptation tasks are given by:\vspace{12pt}

\textbf{Near-Domain Adaptation.} The near-domain adaptation investigates the generalizability of the foundation models to datasets sourced from the same organ but obtained from different institutions than those used in $X_{Meta-train}$ (CRC-TP). This adaptation scenario adopts KBMSC Colon as $X_{Meta-test}$. Specifically, KBSMC Colon contains colorectal tissues, same as CRC-TP, but these tissues were collected and digitized at a different institution. It is noteworthy that CRC-TP is labeled for colorectal tissue sub-typing with 7 classes while KBSMC Colon is labeled for 4-class cancer grading. This adaptation scenario, therefore, assesses the ability of the foundation models to generalize across differences in labeling schemes and institutions, though the tissues were sourced from the same organ.
\vspace{12pt}

\textbf{Middle-Domain Adaptation.} The middle-domain adaptation evaluates the model's ability to generalize to a combination of tissue images from both the same and different organs compared to $X_{Meta-train}$. For this task, LC25000 is used as $X_{Meta-test}$, which involves tissue sample from both the lung and colon. These samples include both benign and carcinoma tissues with 5 different categories. This scenario challenges the foundation models to address variations in labeling schemes and organ and tissue types, testing its robustness in a more complex and heterogeneous domain.
\vspace{12pt}

\textbf{Out-Domain Adaptation.} The out-domain adaptation challenges the foundation models by testing them on tissue images from various organs and institutions that have no overlap with $X_{Meta-train}$. We conduct three out-domain adaptation tasks with three different datasets: PANDA, KBSMC Gastric, and BACH, with each serving as $X_{Meta-test}$. 
These three datasets involve distinct tasks in pathology. Specifically, PANDA is a dataset for 4-class cancer grading, KBSMC Gastric is used for gastric cancer sub-typing with 8 distinct categories, and BACH is for breast cancer detection.  
In this adaptation scenario, we examine the ability of the foundation models  to generalize across entirely new domains, accounting for variations in organ and tissue types, labeling schemes, and institutions.
\vspace{12pt}

We note that, in our experiments, CRC-TP is employed as $X_{Meta-train}$. In contrast, FHist used CRC-TP in a fully supervised setting. The baseline model is replaced by the pathology-specific foundation models in our study, allowing us to explore the effectiveness of the foundation models in adapting to new tasks involving unseen classes during training.

\subsubsection{Few-shot Methods}
\label{subsubsec5}
To adapt the pre-trained pathology foundation models to new tasks, we consider four FSL methods: ProtoNet \citep{snell2017prototypical}, MatchingNet \citep{vinyals2016matching}, \texttt{Baseline} \citep{chen2019closer}, and \texttt{Baseline}++ \citep{chen2019closer}. Given an episode, ProtoNet calculates the prototype (mean) for each class from the feature vectors of the support set, maps the query samples to the prototypes with the closest Euclidean distance, and utilizes this information for fine-tuning the entire model and testing. MatchingNet utilizes the cosine similarity between the support set and the query set. When an episode is given, MatchingNet fine-tunes the entire model and predicts the query samples by weighting the support samples based on their similarity scores. Unlike the previous methods, \texttt{Baseline} and \texttt{Baseline++} fine-tune the model during the testing phase, fixing the feature extractor and repeatedly training only the final linear layer. When an episode is given during the testing phase, \texttt{Baseline} trains a new linear classifier multiple times for that specific episode. Specifically, the linear classifier is trained multiple times on the support set and then used to classify the query samples. \texttt{Baseline++}, an improved version of \texttt{Baseline}, generates class prototypes (weight vectors) by retraining a linear layer multiple times with the support set from the novel classes during the fine-tuning phase. It then predicts the query samples based on the cosine similarity between the query samples and these prototypes. Furthermore, to evaluate the effectiveness of these FSL methods, we also assess the foundation models without fine-tuning by utilizing a K-nearest neighbors (KNN) \citep{guo2003knn} approach.

\subsubsection{Implementation Details}
\label{subsubsec6}
We apply FSL-based adaptation using ProtoNet and MatchingNet to all foundation models, utilizing an $N$-way $K$-shot framework for up to 50,000 iterations on $X_{Meta-train}$, with $N$ set to 4. During the FSL-based adaptation process, we validate the foundation models every 1,000 iterations using 250 randomly sampled episodes from $X_{Meta-val}$, selecting the best-performing models. These selected models are tested on $X_{Test}$ using 1,000 episodes, repeated 1,000 times. \texttt{Baseline} and \texttt{Baseline}++ are performed only during the testing phase. Specifically, using the support samples from each episode of $X_{Test}$, we fix the feature extractor of the pre-trained foundational models and repeatedly fine-tune the final linear layer. This fine-tuning of the last layer is repeated 100 times per episode.
This FSL-based adaptation procedure is repeated for different values of $K$, which are set to 1, 5, and 10, with $Q$ fixed at 15. Hence, the configurations include 4-way-1-shot-15-query, 4-way-5-shot-15-query, and 4-way-10-shot-15-query. All experiments are conducted with Adam optimizer using default parameter values (\( \beta_1 = 0.9 \), \( \beta_2 = 0.999 \), and \( \epsilon = 1.0e-8 \)) and an initial learning rate of 0.05. We adapt color jittering, random horizontal flipping, and random resize cropping data augmentation techniques.
\section{Experiments and Results}
We evaluate the two scenarios used in our experiments by adopting two metrics: 1) Accuracy (Acc) and 2) Macro-average F1 (F1).

\subsection{Consistency Assessment Scenario Experiment}
\label{subsec8}

\begin{figure}[t]
\centering
\includegraphics[height=0.60\textheight, keepaspectratio]{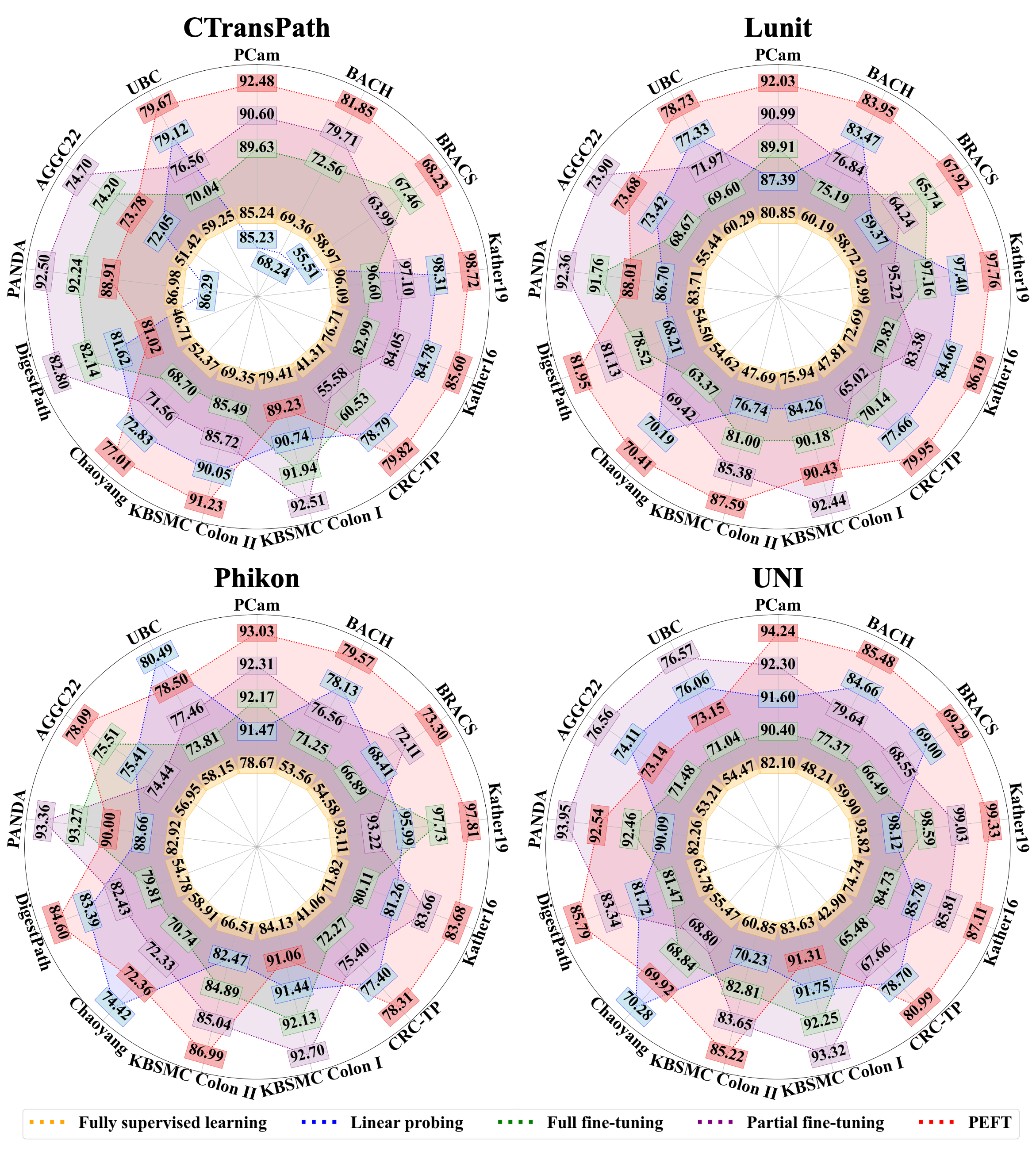}
\caption{Results of consistency assessment scenario.}
\label{fig:benchmark_consistency_result}
\end{figure}

\begin{figure}[]
\centering
\includegraphics[width=\textwidth,keepaspectratio]{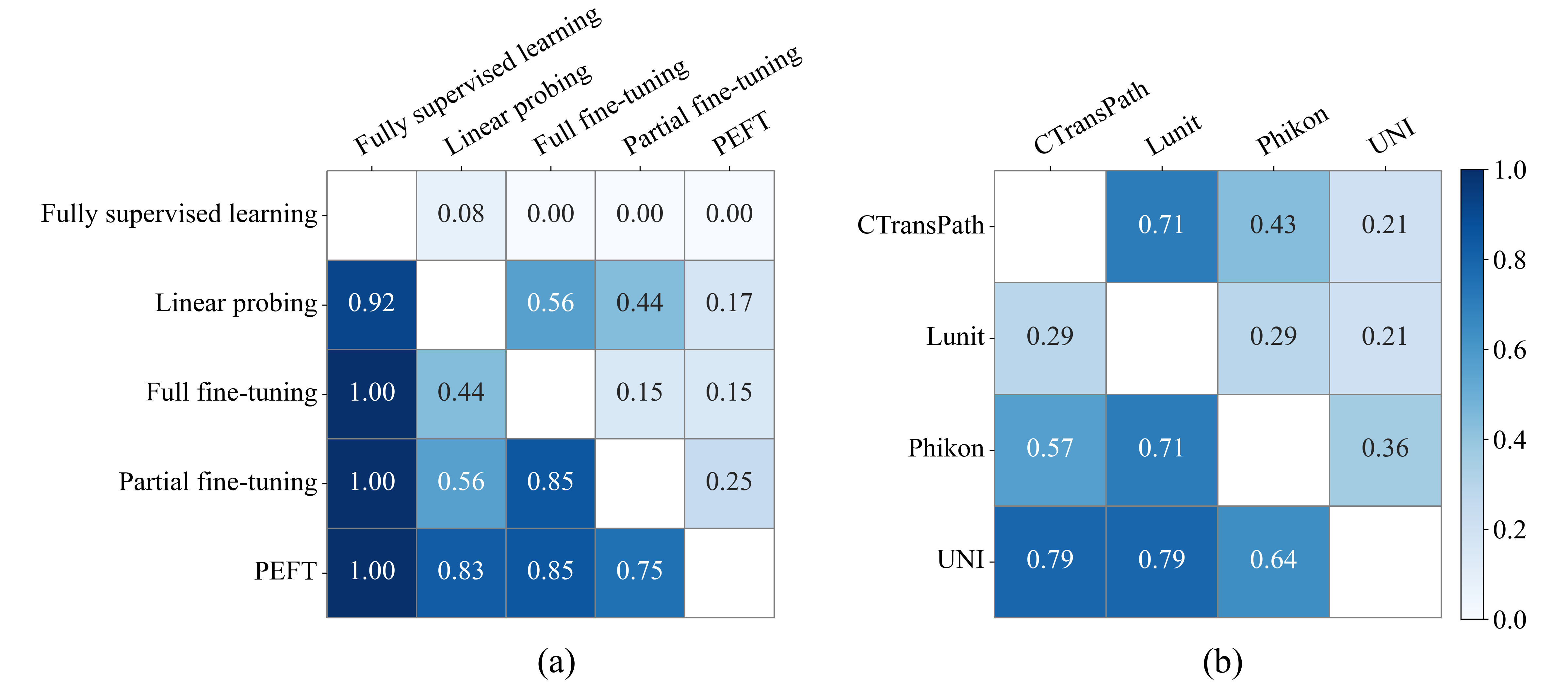}
\caption{Win rate heatmap of (a) five fine-tuning methods and (b) 4 foundation models across 13 datasets.}
\label{fig:win_rate_heatmap_consistency}
\end{figure}

Fig. \ref{fig:benchmark_consistency_result} illustrates the results of the consistency assessment of the four pathology-specific foundation models (CTransPath, Lunit, Phikon, and UNI). In total, 4 classification tasks with 13 datasets were evaluated. Among the five adaptation methods (fully supervised learning, linear probing, full fine-tuning, partial fine-tuning, and PEFT), the strength of PEFT was prominent. It achieved the best performance in 9, 10, 9, and 8 datasets for CTransPath, Lunit, Phikon, and UNI, respectively. Among others, partial fine-tuning was shown to be superior to other three methods (fully supervised learning, linear probing, and full fine-tuning), but its performance varied depending on the datasets. 
In order to provide further insights into the foundation models and their performance, we ranked the four foundation models and five fine-tuning methods using a win rate heatmap, which shows the ratio of the datasets each method obtained the best performance. 
Fig. \ref{fig:win_rate_heatmap_consistency} demonstrates the win rate heatmaps for the five fine-tuning methods and four foundation models. Similar to the observations in Fig. \ref{fig:benchmark_consistency_result}, PEFT provided the highest win rates, substantially outperforming fully supervised by 100\%, linear probing by 83\%, Full fine-tuning by 85\%, and partial fine-tuning by 75\%. This indicates that PEFT is the most suitable and consistent fine-tuning method as compared to other four fine-tuning methods. Among other methods, partial fine-tuning obtained 1.0, 0.56, 0.85, and 0.25 win rate against fully supervised learning, linear probing, full fine-tuning, and PEFT, respectively, suggesting that it is comparable to linear probing and greatly inferior to PEFT. It is also note worthy that linear probing, which is widely used for its simplicity, attained the win rates of 0.92, 0.56, 0.44, and 0.17 over fully supervised learning, full fine-tuning, partial fine-tuning, and PEFT, respectively. These results, particularly the poor performance compared to PEFT, raise questions about the utility of linear probing for the adaptation purposes, since the performance is a primary concern for clinical usage.
Moreover, among the four pathology-specific foundation models, the performance of UNI was striking. It had the highest win rates of 0.79, 0.79, and 0.64 against CTransPath, Lunit, and Phikon, respectively, indicating that UNI is the best performing foundation model within the four tasks and 13 datasets used in this study. Lunit was generally shown to be inferior to other three models. Phikon was slightly better than CTransPath. 

\vspace{12pt}

\subsubsection{Breast Cancer Detection}
\label{subsubsec7}
Table \ref{tab:downstream_breastcancerdetection} shows the results of breast cancer detection using different fine-tuning methods with the four pathology-specific foundation models. 
Among the five fine-tuning methods, PEFT proved to be the most beneficial for the four pathology-specific foundation models (CTransPath, Lunit, Phikon, and UNI). On PCam (fine-tuning dataset), the adoption of PEFT was able to improve Acc and F1 by 1.88$\sim$7.25\% and 0.019$\sim$0.074 for CTransPath, 1.04$\sim$11.18\% and 0.010$\sim$0.114 for Lunit, 0.72$\sim$14.36\% and 0.007$\sim$0.148 for Phikon, and 1.94$\sim$12.14\% and 0.019$\sim$0.123 for UNI, respectively, compared to other four fine-tuning methods. For the other two datasets (BACH and BRACS), we made similar observations. On BACH, PEFT consistently increased Acc and F1 by 2.14$\sim$13.61\% and 0.024$\sim$0.133 for CTransPath, 0.48$\sim$23.76\% and 0.007$\sim$0.237 for Lunit, 1.44$\sim$26.01\% and 0.020$\sim$0.275 for Phikon, and 0.82$\sim$37.27\% and 0.009$\sim$0.405 for UNI, respectively. On BRACS, PEFT also outperformed other four fine-tuning methods regardless of the foundation models except F1 by CTransPath, providing the highest F1 of 0.664.

\subsubsection{Colorectal Cancer Sub-Typing}
\label{subsubsec8}
In colorectal cancer sub-typing, the advantages of PEFT was obvious regardless of the datasets and foundation models (Table \ref{tab:downstream_colonsubtyping}).
On Kather19 (fine-tuning dataset), the combination of PEFT with the found pathology-specific foundation models (CTransPath, Lunit, Phikon, and UNI) achieved the best performance, with Acc of 98.72\%, 97.76\%, 97.81\%, and 99.33\% and F1 of 0.988, 0.978, 0.980, and 0.994, respectively. These were substantially superior to those obtained by the other four fine-tuning methods. 
On Kather16 and CRC-TP, the adoption of PEFT provided similar and consistent improvements. 
Equipped with PEFT, Acc was improved by 0.82$\sim$8.89\% for CTransPath, 1.53$\sim$13.50\% for Lunit, 0.02$\sim$11.86\% for Phikon, and 1.30$\sim$12.37\% for UNI. As for F1, PEFT outperformed other four fine-tuning methods except partial fine-tuning with Phikon, which obtained 0.831 F1.
As the five fine-tuning methods are applied to CRC-TP, the strength of PEFT was remarkable, providing substantail improvements by 1.03$\sim$38.51\% and 0.014$\sim$0.473 for CTransPath, 2.29$\sim$32.14\% and 0.026$\sim$0.383 for Lunit, 0.91$\sim$37.25\% and 0.013$\sim$0.427 for Phikon, and 2.29$\sim$38.09\% and 0.010$\sim$0.509 for UNI, respectively.

\subsubsection{Colorectal Cancer Detection}
\label{subsubsec9}
Table \ref{tab:downstream_coloncancerdetection} demonstrates the results of colorectal cancer detection using the four foundation models and five fine-tuning methods. 
The impact of the five fine-tuning methods varied disproportionately depending on the datasets. On KBSCM Colon I, the application of partial fine-tuning attained the highest Acc and F1 regardless of the foundation models, outperforming that of PEFT by 1.64\%$\sim$3.28\% Acc and 0.016$\sim$0.039 F1. However, partial fine-tuning was not generally effective for other datasets. 
On KBMSC Colon II, PEFT was superior to partial fine-tuning and other fine-tuning methods, consistently enhancing both Acc and F1 across different foundation models such as 1.18$\sim$21.88\% Acc and 0.010$\sim$0.264 F1 for CTransPath, 2.21$\sim$39.9\% Acc and 0.026$\sim$0.391 F1 for Lunit, 1.95$\sim$20.48\% Acc and 0.018$\sim$0.276 F1 for Phikon, and 1.57$\sim$24.37\% Acc and 0.015$\sim$0.239 F1 for UNI, respectively. 
On DigestPath, similar trends were observed. The adoption of PEFT provided a consistent performance gain in F1 by 0.006$\sim$0.324, 0.012$\sim$0.304, 0.029$\sim$0.260, and 0.022$\sim$0.289 for CTransPath, Lunit, Phikon, and UNI, respectively. With respect to Acc, PEFT with Lunit, Phikon, and UNI obtained the best performance of 81.95\%, 84.60\%, and 85.79\% but PEFT with CTransPath was inferior to Linear probing, full fine-tuning, and partial fine-tuning. 
On Chaoyang, PEFT proved particularly effective for CTransPath and Lunit, achieving the highest Acc of 77.01\% and 70.41\% and F1 of 0.768 and 0.704, respectively. However, for Phikon and UNI, PEFT was inferior to linear probing, which obtained Acc of 74.42\% and 70.28\% and F1 of 0.743 and 0.701, respectively.

\subsubsection{Prostate Cancer Grading}
\label{subsubsec10}
The results of prostate cancer grading on three test datasets are available in Table \ref{tab:downstream_prostatecancergrading}. 
On PANDA (fine-tuning dataset), partial fine-tuning showed the highest performance across different foundation models, achieving 92.36\%$\sim$92.50\% Acc and 0.898$\sim$0.918 F1. 
On the two out-of-domain datasets (AGGC22 and UBC), the results varied depending on the specific combination of fine-tuning methods and foundation models and the choice of evaluation metrics. 
On AGGC22, the usage of partial fine-tuning resulted in the highest Acc of 74.70\%, 73.90\%, and 76.56\% for CTransPath, Lunit, and UNI, respectively. However, these obtained subtantially poorer F1 compared to those obtained using PEFT. In terms of F1, linear probing with Lunit and UNI achieved the highest scores of 0.608 and 0.607, respectively, and PEFT with CTransPath and Phikon produced the best scores, reaching 0.602 and 0.629, respectively. 
For Phikon, PEFT obtained the highest Acc and F1 of 78.09\% and 0.629, respectively. Notably, no other combination of fine-tuning methods and foundation models was able to attain the highest Acc and F1 simultaneously.
On UBC, partial-fine tuning was inferior to either linear probing or PEFT except Acc for UNI. Linear probing managed to achieve the highest F1 of 0.616 and 0.621 for CTransPath and UNI, respectively, and the highest Acc of 80.49\% for Phikon. PEFT produced the best results for CTransPath and Phikon, with the top Acc and F1 of 79.67\% 0.622, respectively. It is remarkable that, for Lunit, PEFT achieved both the highest Acc (78.73\%) and F1 (0.625), making it the only combination to attain the best results for both evaluation metrics at the same time. \vspace{12pt}

\subsubsection{Computational Complexity for Consistency Assessment Scenario}
\label{subsec_comp}
We measured and compared the computational complexity of different fine-tuning methods across four pathology-specific foundation models. For each combination, we computed the number of parameters and measured the average execution time and the peak memory usage during both training and testing per image and batch. We set the batch size to 64. The results are available in Table \ref{tab:model_complexity}. 
Among the four foundation models, UNI has the highest number of parameters \( >303.0 \text{ million} \), while CTransPath and Lunit require a relatively smaller number of parameters \( <303.0 \text{ million} \). 
For these foundation models, usage of linear probing, partial fine-tuning, and full fine-tuning did not add any additional parameters, whereas PEFT increased the number of parameters by approximately 1\%. 
However, the memory requirement varied significantly depending on the fine-tuning methods, particularly during training. For instance, linear probing consumed the least memory, as it primarily adjusts only the final layer without modifying additional layers. In contrast, full fine-tuning modifies all layers, leading to the highest memory consumption, which is 3 to 13 times greater than linear probing, depending on the foundation models. 
During testing, the memory usage decreased as all parameters remained fixed, resulting in comparable memory requirements across different fine-tuning methods. PEFT caused only a negligible increase in memory usage due to the increase in the number of parameters. 
With respect to the execution time, larger models generally required more time to execute, especially during training. Among the fine-tuning methods, full fine-tuning was typically the slowest during training, while linear probing was the fastest. However, during testing, the execution time across the fine-tuning methods was more or less the same within each foundation model.

\vspace{12pt}

\begin{table}[]
\caption{Fine-tuning adaptation results on breast cancer detection.}
\label{tab:downstream_breastcancerdetection}
\begin{adjustbox}{max width=\textwidth}
\begin{tabular}{@{}llcccccccccc@{}}
\toprule
\multicolumn{1}{c}{\multirow{2}{*}{\parbox{3.5cm}{Test Dataset}}} & \multicolumn{1}{c}{\multirow{2}{*}{Model}} & \multicolumn{2}{c}{Fully supervised learning} & \multicolumn{2}{c}{Linear probing} & \multicolumn{2}{c}{Full fine-tuning} & \multicolumn{2}{c}{Partial fine-tuning} & \multicolumn{2}{c}{PEFT} \\ \cmidrule(l){3-12}
                              &             & Acc (\%)  & F1        & Acc (\%)  & F1        & Acc (\%)  & F1        & Acc (\%)  & F1        & Acc (\%)  & F1        \\ \midrule
PCam                          & CTransPath  & 85.24     & 0.851     & 85.23     & 0.852     & 89.63     & 0.896     & 90.60     & 0.906     & {\textbf{92.48}}     & {\textbf{0.925}}     \\
                              & Lunit       & 80.85     & 0.806     & 87.39     & 0.874     & 89.91     & 0.899     & 90.99     & 0.910     & {\textbf{92.03}}     & {\textbf{0.920}}     \\
                              & Phikon      & 78.67     & 0.782     & 91.47     & 0.914     & 92.17     & 0.921     & 92.31     & 0.923     & {\textbf{93.03}}     & {\textbf{0.930}}     \\
                              & UNI         & 82.10     & 0.819     & 91.60     & 0.916     & 90.40     & 0.904     & 92.30     & 0.923     & {\textbf{94.24}}     & {\textbf{0.942}}     \\ \midrule
BACH                          & CTransPath  & 69.36     & 0.679     & 68.24     & 0.677     & 72.56     & 0.684     & 79.71     & 0.786     & {\textbf{81.85}}     & {\textbf{0.810}}     \\
                              & Lunit       & 60.19     & 0.593     & 83.47     & 0.823     & 75.19     & 0.731     & 76.84     & 0.751     & {\textbf{83.95}}     & {\textbf{0.830}}     \\
                              & Phikon      & 53.56     & 0.508     & 78.13     & 0.763     & 71.25     & 0.661     & 76.56     & 0.738     & {\textbf{79.57}}     & {\textbf{0.783}}     \\
                              & UNI         & 48.21     & 0.444     & 84.66     & 0.840     & 77.37     & 0.765     & 79.64     & 0.787     & {\textbf{85.48}}     & {\textbf{0.849}}     \\ \midrule
BRACS                         & CTransPath  & 58.97     & 0.579     & 55.51     & 0.490     & 67.46     & \textbf{0.664}     & 63.99     & 0.607     & {\textbf{68.23}}     & 0.647     \\
                              & Lunit       & 58.72     & 0.568     & 59.37     & 0.494     & 65.74     & 0.637     & 64.24     & 0.596     & {\textbf{67.92}}     & {\textbf{0.645}}     \\
                              & Phikon      & 54.58     & 0.533     & 68.41     & 0.651     & 66.89     & 0.654     & 72.11     & 0.708     & {\textbf{73.30}}     & {\textbf{0.720}}     \\
                              & UNI         & 59.90     & 0.593     & 69.00     & 0.651     & 66.49     & 0.646     & 68.55     & 0.653     & {\textbf{69.29}}     & {\textbf{0.662}}     \\ \bottomrule
\end{tabular}
\end{adjustbox}
\end{table}

\begin{table}[]
\caption{Fine-tuning adaptation results on colorectal cancer sub-typing.}
\label{tab:downstream_coloncancerdetection}
\begin{adjustbox}{max width=\textwidth}
\begin{tabular}{@{}llcccccccccc@{}}
\toprule
\multicolumn{1}{c}{\multirow{2}{*}{\parbox{3.5cm}{Test Dataset}}} & \multicolumn{1}{c}{\multirow{2}{*}{Model}} & \multicolumn{2}{c}{Fully supervised learning} & \multicolumn{2}{c}{Linear probing} & \multicolumn{2}{c}{Full fine-tuning} & \multicolumn{2}{c}{Partial fine-tuning} & \multicolumn{2}{c}{PEFT} \\ \cmidrule(l){3-12}
                              &             & Acc (\%)  & F1        & Acc (\%)  & F1        & Acc (\%)  & F1        & Acc (\%)  & F1        & Acc (\%)  & F1        \\ \midrule
Kahter19                      & CTransPath  & 96.09     & 0.960     & 98.31     & 0.983     & 96.60     & 0.969     & 97.10     & 0.973     & {\textbf{98.72}}     & {\textbf{0.988}}     \\
                              & Lunit       & 92.99     & 0.931     & 97.40     & 0.975     & 97.16     & 0.973     & 95.22     & 0.957     & {\textbf{97.76}}     & {\textbf{0.978}}     \\
                              & Phikon      & 93.11     & 0.933     & 95.99     & 0.965     & 97.73     & 0.979     & 93.22     & 0.939     & {\textbf{97.81}}     & {\textbf{0.980}}     \\
                              & UNI         & 93.82     & 0.936     & 98.12     & 0.983     & 98.59     & 0.986     & 99.03     & 0.991     & {\textbf{99.33}}     & {\textbf{0.994}}     \\ \midrule
Kahter16                      & CTransPath  & 76.71     & 0.756     & 84.78     & 0.849     & 82.99     & 0.823     & 84.05     & 0.838     & {\textbf{85.60}}     & {\textbf{0.855}}     \\
                              & Lunit       & 72.69     & 0.716     & 84.66     & 0.843     & 79.82     & 0.788     & 83.38     & 0.831     & {\textbf{86.19}}     & {\textbf{0.863}}     \\
                              & Phikon      & 71.82     & 0.707     & 81.26     & 0.802     & 80.11     & 0.792     & 83.66     & {\textbf{0.831}}     & {\textbf{83.68}}     &  0.828    \\
                              & UNI         & 74.74     & 0.731     & 85.78     & 0.854     & 84.73     & 0.843     & 85.81     & 0.855     & {\textbf{87.11}}     & {\textbf{0.869}}     \\ \midrule
CRC-TP                        & CTransPath  & 41.31     & 0.249     & 78.79     & 0.708     & 60.53     & 0.485     & 55.58     & 0.378     & {\textbf{79.82}}     & {\textbf{0.722}}     \\
                              & Lunit       & 47.81     & 0.328     & 77.66     & 0.685     & 70.14     & 0.538     & 65.02     & 0.509     & {\textbf{79.95}}     & {\textbf{0.711}}     \\
                              & Phikon      & 41.06     & 0.278     & 77.40     & 0.692     & 72.27     & 0.604     & 75.40     & 0.652     & {\textbf{78.31}}     & {\textbf{0.705}}     \\
                              & UNI         & 42.90     & 0.219     & 78.70     & 0.718     & 65.48     & 0.522     & 67.66     & 0.557     & {\textbf{80.99}}     & {\textbf{0.728}}     \\ \bottomrule
\end{tabular}
\end{adjustbox}
\end{table}

\begin{table}[]
\caption{Fine-tuning adaptation results on colorectal cancer detection.}
\label{tab:downstream_colonsubtyping}
\begin{adjustbox}{max width=\textwidth}
\begin{tabular}{@{}llcccccccccc@{}}
\toprule
\multicolumn{1}{c}{\multirow{2}{*}{\parbox{3.5cm}{Test Dataset}}} & \multicolumn{1}{c}{\multirow{2}{*}{Model}} & \multicolumn{2}{c}{Fully supervised learning} & \multicolumn{2}{c}{Linear probing} & \multicolumn{2}{c}{Full fine-tuning} & \multicolumn{2}{c}{Partial fine-tuning} & \multicolumn{2}{c}{PEFT} \\ \cmidrule(l){3-12}
                              &             & Acc (\%)  & F1        & Acc (\%)  & F1        & Acc (\%)  & F1        & Acc (\%)  & F1        & Acc (\%)  & F1        \\ \midrule
KBSMC Colon I                 & CTransPath  & 79.41     & 0.776     & 90.74     & 0.902     & 91.94     & 0.917     & {\textbf{92.51}}     & {\textbf{0.922}}     & 89.23     & 0.883     \\
                              & Lunit       & 75.94     & 0.736     & 84.26     & 0.835     & 90.18     & 0.898     & {\textbf{92.44}}     & {\textbf{0.921}}     & 90.43     & 0.901     \\
                              & Phikon      & 84.13     & 0.832     & 91.44     & 0.911     & 92.13     & 0.919     & {\textbf{92.70}}     & {\textbf{0.923}}     & 91.06     & 0.907     \\
                              & UNI         & 83.63     & 0.833     & 91.75     & 0.913     & 92.25     & 0.920     & {\textbf{93.32}}     & {\textbf{0.930}}     & 91.31     & 0.910     \\ \midrule
KBSMC Colon II                & CTransPath  & 69.35     & 0.636     & 90.05     & 0.890     & 85.49     & 0.846     & 85.72     & 0.849     & {\textbf{91.23}}     & {\textbf{0.900}}     \\
                              & Lunit       & 47.69     & 0.476     & 76.74     & 0.736     & 81.00     & 0.803     & 85.38     & 0.841     & {\textbf{87.59}}     & {\textbf{0.867}}     \\
                              & Phikon      & 66.51     & 0.584     & 82.47     & 0.818     & 84.89     & 0.840     & 85.04     & 0.842     & {\textbf{86.99}}     & {\textbf{0.860}}     \\
                              & UNI         & 60.85     & 0.603     & 70.23     & 0.700     & 82.81     & 0.819     & 83.65     & 0.827     & {\textbf{85.22}}     & {\textbf{0.842}}     \\ \midrule
Chaoyang                      & CTransPath  & 52.37     & 0.524     & 72.83     & 0.727     & 68.70     & 0.687     & 71.56     & 0.716     & {\textbf{77.01}}     & {\textbf{0.768}}     \\
                              & Lunit       & 54.62     & 0.546     & 70.19     & 0.684     & 63.37     & 0.633     & 69.42     & 0.694     & {\textbf{70.41}}     & {\textbf{0.704}}     \\
                              & Phikon      & 58.91     & 0.581     & {\textbf{74.42}}     & {\textbf{0.743}}     & 70.74     & 0.707     & 72.33     & 0.723     & 72.36     & 0.722     \\
                              & UNI         & 55.47     & 0.554     & {\textbf{70.28}}     & {\textbf{0.701}}     & 68.84     & 0.688     & 68.80     & 0.688     & 69.92     & 0.699     \\ \midrule
DigestPath                    & CTransPath  & 46.71	    & 0.448	    & 81.62	    & 0.766	    & 82.14	    & 0.743	    & {\textbf{82.80}}	    & 0.761	    & 81.02	   & {\textbf{0.772}}     \\    
                              & Lunit       & 54.50	  & 0.449	  & 68.21	  & 0.648	  & 78.52	  & 0.694	  & 81.13	  & 0.741	  & {\textbf{81.95}}	 & {\textbf{0.753}}     \\
                              & Phikon      & 54.78	  & 0.516	  & 83.39	  & 0.747	  & 79.81	  & 0.705	  & 82.43	  & 0.734	  & {\textbf{84.60}}	 & {\textbf{0.776}}     \\
                              & UNI         & 63.78	  & 0.510	  & 81.72	  & 0.752	  & 81.47	  & 0.739	  & 83.34	  & 0.777	  & {\textbf{85.79}}	 & {\textbf{0.799}}     \\ \bottomrule
\end{tabular}
\end{adjustbox}
\end{table}

\begin{table}[]
\caption{Fine-tuning adaptation results on prostate cancer grading.}
\label{tab:downstream_prostatecancergrading}
\begin{adjustbox}{max width=\textwidth}
\begin{tabular}{@{}llcccccccccc@{}}
\toprule
\multicolumn{1}{c}{\multirow{2}{*}{\parbox{3.5cm}{Test Dataset}}} & \multicolumn{1}{c}{\multirow{2}{*}{Model}} & \multicolumn{2}{c}{Fully supervised learning} & \multicolumn{2}{c}{Linear probing} & \multicolumn{2}{c}{Full fine-tuning} & \multicolumn{2}{c}{Partial fine-tuning} & \multicolumn{2}{c}{PEFT} \\ \cmidrule(l){3-12}
                              &             & Acc (\%)  & F1        & Acc (\%)  & F1        & Acc (\%)  & F1        & Acc (\%)  & F1        & Acc (\%)  & F1        \\ \midrule
PANDA                         & CTransPath  & 86.98     & 0.828     & 86.29     & 0.817     & 92.24     & 0.895     & {\textbf{92.50}}     & {\textbf{0.898}}     & 88.91     & 0.845  \\
                              & Lunit       & 83.71     & 0.797     & 86.70     & 0.829     & 91.76     & 0.890     & {\textbf{92.36}}     & {\textbf{0.898}}     & 88.01     & 0.843  \\
                              & Phikon      & 82.92     & 0.785     & 88.66     & 0.850     & 93.27     & 0.911     & {\textbf{93.36}}     & {\textbf{0.911}}     & 90.00     & 0.866  \\
                              & UNI         & 82.26     & 0.775     & 90.09     & 0.868     & 92.46     & 0.899     & {\textbf{93.95}}     & {\textbf{0.918}}     & 92.54     & 0.902  \\ \midrule
AGGC22                        & CTransPath  & 51.42     & 0.361     & 72.05     & 0.569     & 74.20     & 0.592     & {\textbf{74.70}}     & 0.552     & 73.78     & {\textbf{0.602}}  \\
                              & Lunit       & 55.44     & 0.317     & 73.42     & {\textbf{0.608}}     & 68.67     & 0.494     & {\textbf{73.90}}     & 0.547     & 73.68     & 0.600  \\
                              & Phikon      & 56.95     & 0.349     & 75.41     & 0.575     & 75.51     & 0.579     & 74.44     & 0.572     & {\textbf{78.09}}     & {\textbf{0.629}}  \\
                              & UNI         & 53.21     & 0.346     & 74.11     & {\textbf{0.607}}     & 71.48     & 0.519     & {\textbf{76.56}}     & 0.585     & 73.14     & 0.600  \\ \midrule
UBC                           & CTransPath  & 59.25     & 0.432     & 79.12     & {\textbf{0.616}}     & 70.04     & 0.524     & 76.56     & 0.583     & {\textbf{79.67}}     & 0.605  \\
                              & Lunit       & 60.29     & 0.412     & 77.33     & 0.614     & 69.60     & 0.504     & 71.97     & 0.543     & {\textbf{78.73}}     & {\textbf{0.625}}  \\
                              & Phikon      & 58.15     & 0.433     & {\textbf{80.49}}     & 0.618     & 73.81     & 0.569     & 77.46     & 0.597     & 78.50     & {\textbf{0.622}}  \\
                              & UNI         & 54.47     & 0.426     & 76.06     & {\textbf{0.621}}     & 71.04     & 0.504     & {\textbf{76.57}}     & 0.568     & 73.15     & 0.606  \\ \bottomrule
\end{tabular}
\end{adjustbox}
\end{table}

\begin{table}[]
\caption{Computational complexity of foundation models and fine-tuning methods.}
\label{tab:model_complexity}
\centering
\begin{adjustbox}{max width=\textwidth}
\begin{tabular}{@{}llccccccccc@{}}
\toprule
\multirow{2}{*}{Model}      & \multirow{2}{*}{Method} & \multicolumn{1}{l}{\multirow{2}{*}{\# Params (M)}} & \multicolumn{2}{l}{Memory (MB/batch)} & \multicolumn{2}{l}{Memory (MB/image)} & \multicolumn{2}{l}{Time (ms/batch)} & \multicolumn{2}{l}{Time (ms/image)} \\ \cmidrule(l){4-11} 
                            &                         & \multicolumn{1}{l}{} & Train              & Test             & Train              & Test              & Train             & Test            & Train            & Test              \\ \midrule
\multirow{4}{*}{CTrasnPath} & Linear probing          & 27.523                & 2,363              & 2,355            & 733                & 595               & 82.59             & 81.55           & 18.83            & 15.02             \\
                            & Partial fine-tuning     & 27.523                & 2,961              & 2,355            & 973                & 595               & 128.76            & 81.55           & 34.84            & 14.97             \\
                            & Full fine-tuning        & 27.523                & 7,949              & 2,355            & 1,083              & 595               & 225.74            & 81.50           & 48.04            & 15.03             \\
                            & PEFT                    & 27.804                & 6,017              & 2,375            & 777                & 617               & 197.18            & 81.40           & 40.82            & 15.10             \\ \midrule
\multirow{4}{*}{Lunit}      & Linear probing          & 21.667                & 885                & 879              & 451                & 443               & 53.39             & 54.41           & 12.71            & 10.49             \\
                            & Partial fine-tuning     & 21.667                & 3,341              & 879              & 687                & 443               & 109.38            & 54.27           & 24.15            & 10.78             \\
                            & Full fine-tuning        & 21.667                & 5,243              & 879              & 847                & 443               & 150.16            & 54.29           & 35.00            & 10.70             \\
                            & PEFT                    & 21.814                & 4,115              & 884              & 545                & 456               & 143.02            & 54.30           & 28.70            & 10.53             \\ \midrule
\multirow{4}{*}{Phikon}     & Linear probing          & 85.802                & 1,549              & 1,541            & 745                & 717               & 168.64            & 170.71          & 13.88            & 11.93             \\
                            & Partial fine-tuning     & 85.802                & 7,555              & 1,541            & 1,729              & 717               & 354.58            & 171.10          & 36.16            & 11.81             \\
                            & Full fine-tuning        & 85.802                & 10,729             & 1,541            & 2,263              & 717               & 449.72            & 171.07          & 42.42            & 11.98             \\
                            & PEFT                    & 86.097                & 7,857              & 1,552            & 823                & 726               & 357.55            & 175.39          & 37.17            & 11.96             \\ \midrule
\multirow{4}{*}{UNI}        & Linear probing          & 303.355              & 2,331              & 2,323            & 1,545              & 1,517             & 523.36            & 526.08          & 17.66            & 15.56             \\
                            & Partial fine-tuning     & 303.355              & 21,555             & 2,323            & 4,451              & 1,517             & 1265.44           & 526.92          & 61.21            & 15.40             \\
                            & Full fine-tuning        & 303.355              & 27,039             & 2,323            & 6,483              & 1,517             & 1424.93           & 526.30          & 96.91            & 15.70             \\
                            & PEFT                    & 303.549               & 15,561             & 2,332            & 2,061              & 1,533             & 1383.58           & 525.72          & 66.73            & 15.72             \\ \bottomrule
\end{tabular}
\end{adjustbox}
\end{table}

\begin{figure}[ht]
\centering
\includegraphics[width=\textwidth,keepaspectratio]{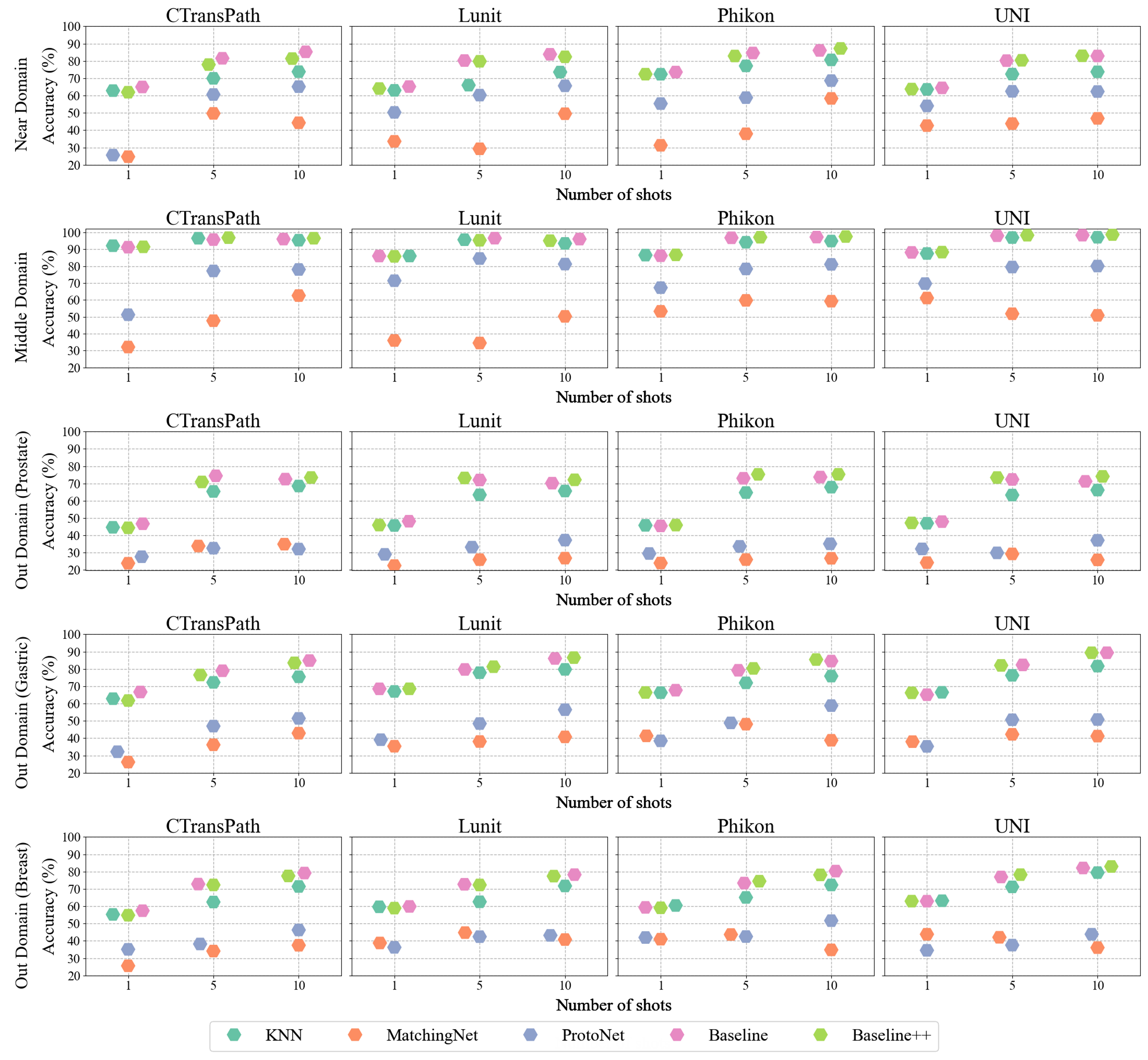}
\caption{Results of flexibility assessment scenario.}
\label{fig:fewshotresult}
\end{figure}

\begin{figure}[]
\centering
\includegraphics[width=\textwidth,keepaspectratio]{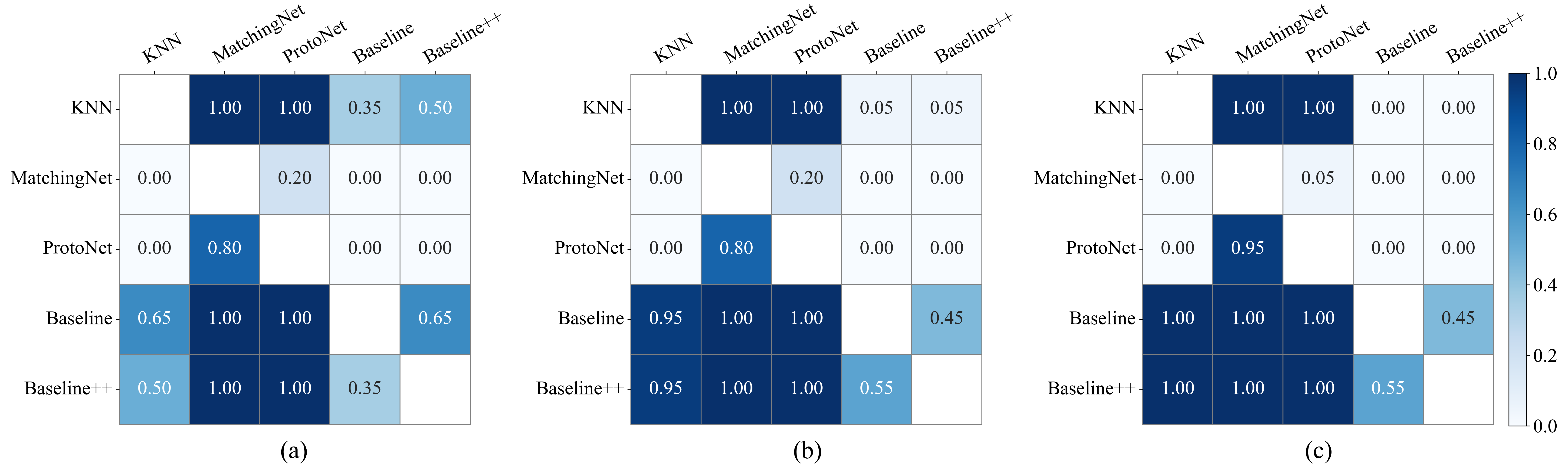}
\caption{Win rate heatmap of (a) 1-shot, (b) 5-shot, and (c) 10-shot few-shot learning methods across 5 datasets.}
\label{fig:win_rate_heatmap_flexibiltiy}
\end{figure}

\subsection{Flexibility Assessment Scenario Experiment}
\label{subsec9}

Fig. \ref{fig:fewshotresult} demonstrates the results of flexibility assessment over five FSL methods (KNN, MatchingNet, ProtoNet, \texttt{Baseline}, and \texttt{Baseline}++) with shot counts of 1, 5, and 10. Three adaptation tasks, including near-domain, middle-domain, and out-domain adaptation tasks, were evaluated. The out-domain adaptation task contains three sub-tasks such as prostate cancer grading, gastric cancer sub-typing, and breast tissue sub-typing. Within the five FSL methods, \texttt{Baseline} and \texttt{Baseline}++ generally stood out as the top-performing methods. MatchingNet and ProtoNet were found to be substantially less effective in the adaptation tasks compared to other methods. KNN was comparable to \texttt{Baseline} and \texttt{Baseline}++, particularly in the 1-shot scenario. 

Moreover, we recorded the ranking of the five FSL methods over the three adaptation tasks and computed the win rates. 
Fig. \ref{fig:win_rate_heatmap_flexibiltiy} shows the win rate heatmaps for the five FSL methods in the 1-shot, 5-shot, and 10-shot scenarios. 
In the 1-shot scenario, \texttt{Baseline} produced the highest win rates, obtaining 0.65 against KNN, 1.00 against both MatchingNet and ProtoNet, and 0.65 against \texttt{Baseline}++. 
Both \texttt{Baseline}++ and KNN were also dominant compared to MatchingNet and ProtoNet, each achieving win rates of 1.00 against these two models. \texttt{Baseline}++ and KNN were comparable to each other.
Regarding the 5-shot and 10-shot scenarios, the dominance of \texttt{Baseline} and \texttt{Baseline}++ was further highlighted with the win rates of 0.95, 1.00, and 1.00 over KNN, MatchingNet, and ProtoNet. \texttt{Baseline}++ was slightly better than \texttt{Baseline} by the win rate of 0.55. Though KNN surpassed MatchingNet and ProtoNet, it was clearly inferior to both \texttt{Baseline} and \texttt{Baseline}++.

\subsubsection{Near-Domain Adaptation}
\label{subsubsec11}

\sloppy
The results of the near-domain adaptation are available in Table \ref{tab:fewshot_near_middle}. \texttt{Baseline} generally achieved the best results for both Acc and F1 regardless of the foundation models, evaluation metrics, and number of shots, with the exception of Phikon with 10 shots and UNI with 5 and 10 shots. For these three cases, \texttt{Baseline} was ranked as the second-best model. As for other cases, \texttt{Baseline}++ was typically found to be the second-best model except CTransPath with 1 shot. 
In a head-to-head comparison among the four pathology-specific foundation models, Phikon with \texttt{Baseline} achieved the highest Acc and F1 across different number of shots, substantially outperforming other three models by 8.37\%$\sim$9.13\% Acc and 0.090$\sim$0.100 F1 for 1 shot, 3.05\%$\sim$4.41\% Acc and 0.032$\sim$0.040 F1 for 5 shots, and 0.80\%$\sim$3.20\% Acc and 0.007$\sim$0.030 F1 for 10 shots. 
In terms of shot counts, Phikon with \texttt{Baseline} produced the best results for 1 (73.56\% Acc and 0.727 F1) and 5 shots (84.57\% Acc and 0.845 F1). However, for 10 shots, Phikon with \texttt{Baseline}++ attained the highest Acc and F1 of 87.21\% and 0.871, respectively.


\subsubsection{Middle-Domain Adaptation}
\label{subsubsec12}

Table \ref{tab:fewshot_near_middle} shows the results of the middle-domain adaptation. For CtransPath, Phikon and UNI, \texttt{Baseline}++ was generally superior to other adaptation methods across different shot counts and evaluation metrics except for CTransPath with 1 shot. 
As for Lunit, while KNN outperformed others using 1 shot, \texttt{Baseline} produced the top performance using 5 shots (96.63\% Acc and 0.966 F1) and 10 shots (96.08\% Acc and 0.960 F1). 
With respect to the shot counts and foundation models, equipped with KNN, CTransPath achieved the best Acc of 92.15\% and F1 of 0.919 with 1 shot, greatly outperforming other foundation models by 3.79\%$\sim$6.09\% Acc and 0.045$\sim$0.070 F1 as compared to the best results obtained by each model with 1 shot.
Using 5 and 10 shots, UNI with \texttt{Baseline}++ attained the top results, with 98.37\% Acc and 0.984 F1 for 5 shots and 98.76\% Acc and 0.988 F1 for 10 shots. This presents an improvement of 1.19\%$\sim$1.74\% Acc and 0.012$\sim$0.018 F1 for 5 shots and an improvement of 1.15\%$\sim$2.68\% Acc and 0.012$\sim$0.028 F1 for 10 shots over the best-performing combinations from other foundation models.

\subsubsection{Out-Domain Adaptation: Prostate Cancer Grading}
\label{subsubsec13}

In the out-domain adaptation to prostate cancer grading (Table \ref{tab:fewshot_out}), \texttt{Baseline} and \texttt{Baseline}++ were shown to be the top-performing models across various shot counts and foundation models. 
Specifically, for CTransPath, \texttt{Baseline} obtained the highest results for the 1 shot (46.74\% Acc and 0.439 F1) and 5 shots (74.44\% Acc and 0.737 F1), meanwhile  \texttt{Baseline}++ achieved the highest Acc of 73.46\% and F1 of 0.728. 
As for other foundation models (Phikon, Lunit, and UNI), \texttt{Baseline}++ outperformed other adaptation methods, particularly effective in scenarios with 5 shots and 10 shots. In the scenario with 1 shot, \texttt{Baseline} was beneficial for Lunit and UNI, surpassing other methods. As for Phikon, \texttt{Baseline}++ obtained the highest Acc of 45.94\% and the second-highest F1 of 0.427, with KNN achieving the highest F1 of 0.435. 

\subsubsection{Out-Domain Adaptation: Gastric Cancer Sub-typing}
\label{subsubsec14}
Similarly, \texttt{Baseline} and \texttt{Baseline}++ demonstrated effectiveness in the out-domain adaptation for gastric cancer sub-typing. (Table \ref{tab:fewshot_out}).
For CTransPath, \texttt{Baseline} was consistently superior to other methods regardless of shot counts and evaluation metrics. 
In contrast, for Lunit, \texttt{Baseline}++ delivered the best results for every shot count and evaluation metric. 
With Phikon, \texttt{Baseline} proved to be the most advantageous for the 1 shot scenario, and, in the scenarios with 5 and 10 shots, \texttt{Baseline}++ outperformed other methods. 
When it comes to UNI, KNN with 1 shot obtained the best performance, while \texttt{Baseline}++ achieved the highest Acc and F1 for both 5 and 10 shots.
Regarding the combinations of the foundation models and shot counts, Lunit with \texttt{Baseline}++ attained the best results for the 1-shot scenario, with 68.52\% Acc and 0.680 F1. For the scenarios with 5 and 10 shots, the combination of UNI and \texttt{Baseline} obtained the highest scores, with 82.32\% Acc and 0.824 F1 for 5 shots and 89.34\% Acc and 0.894 F1 for 10 shots.

\subsubsection{Out-Domain Adaptation: Breast Cancer Detection}
\label{subsubsec15}
The results of the out-domain adaptation for breast cancer detection further confirmed the strength of \texttt{Baseline} and \texttt{Baseline}++, as shown in Table \ref{tab:fewshot_out}. 
For both CTransPath and Lunit, \texttt{Baseline} constantly proved to be the best adaptation method across different shot counts and evaluation metrics. The second-best methods varied; KNN ranked second for the 1 shot scenario while \texttt{Baseline}++ took the second place in the 5 and 10 shot scenarios.
With regard to Phikon and UNI, the results varied depending on shot counts and foundation models. 
For both Phikon and UNI, KNN was the top performer in the 1-shot scenario and \texttt{Baseline}++ was the best-performing method in the 5-shot scenario. With 10 shots, \texttt{Baseline} and \texttt{Baseline}++ obtained the superior performance for Phikon and UNI, respectively. 
Comparing the four pathology-specific foundation models, UNI proved to be the most effective adaptation method. Specifically, using 1 shot, UNI with KNN delivered the best results of 63.16\% Acc and 0.622 F1. For 5 and 10 shots, UNI, paired with \texttt{Baseline}++, found to be the top-performing model, achieving an Acc of 78.22\% and F1 of 0.774 F1 for 5 shots and an Acc of 82.94\% and F1 of 0.830 for 10 shots.
\vspace{12pt}

\vspace{12pt}

\subsubsection{Computational Complexity for Flexibility Assessment Scenario}
\label{subsec_comp_flex}
Moreover, we evaluated the computational complexity of the four pathology-specific foundation models using various FSL methods. 
For each combination, we measured the number of parameters, the average execution time, and peak memory usage during training and testing per episode for 1-shot, 5-shot, and 10-shot settings (Table \ref{tab:model_complexity_fewshot}). 
Among the FSL methods, KNN, ProtoNet, and MatchingNet did not require any additional parameters, whereas \texttt{Baseline} and \texttt{Baseline}++ resulted in a slight increase in the number of parameters, accounting for less than 0.1\% of the original model size. 
Memory requirements varied depending on both the foundation models and FSL methods. Larger foundation models consumed more memory, and the FSL methods further contribute to the overall memory usage.
For instance, ProtoNet and MatchingNet adjust all layers in the foundation models, leading to the highest memory consumption. In contrast, KNN, \texttt{Baseline}, and \texttt{Baseline}++ do not consume memory during training, as KNN, \texttt{Baseline}, and \texttt{Baseline}++ do not involve any training. 
During testing, since all parameters remained fixed, ProtoNet and MatchingNet demonstrate substantially reduced memory usage compared to the training phase for all four foundation models. Specifically, there was a 2.5 to 3.5 fold decrease for CTransPath, a 3.0 to 4.0 fold decrease for Lunit, a 4.3 to 5.8 fold decrease for Phikon, and a 7.8 to 10.8 fold decrease for UNI across different shot counts. Meanwhile, \texttt{Baseline} and \texttt{Baseline}++ exhibited slightly increased memory usage compared to ProtoNet, MatchingNet, and KNN due to the modifications of the final linear layer. However, the execution time for the foundation models using \texttt{Baseline} and \texttt{Baseline}++ was greatly slower compared to other FSL methods due to their design. With ProtoNet, MatchingNet, and KNN, the foundation models required less than 2.90 seconds to process one episode, whereas they took more than 202.00 seconds when using \texttt{Baseline} and \texttt{Baseline}++.

\vspace{12pt}

\begin{table}[]
\caption{Few-shot learning results on near-domain and middle-domain adaptation tasks.}
\label{tab:fewshot_near_middle}
\centering
\begin{adjustbox}{max width=\textwidth}
\begin{tabular}{@{}llcccccc|cccccc@{}}
\toprule
           &                     & \multicolumn{6}{c|}{Near-domain (Colorectal cancer grading)}                                                                         & \multicolumn{6}{c}{Middle-domain (Lung and colon sub-typing)}                                                                      \\ \midrule
           &                     & \multicolumn{2}{c}{1-shot} & \multicolumn{2}{c}{5-shot} & \multicolumn{2}{c|}{10-shot} & \multicolumn{2}{c}{1-shot} & \multicolumn{2}{c}{5-shot} & \multicolumn{2}{c}{10-shot} \\ \midrule
Model      & Method              & \begin{tabular}[c]{@{}c@{}}Acc\\ (\%)\end{tabular}        & F1             & \begin{tabular}[c]{@{}c@{}}Acc\\ (\%)\end{tabular}        & F1             & \begin{tabular}[c]{@{}c@{}}Acc\\ (\%)\end{tabular}         & F1              & \begin{tabular}[c]{@{}c@{}}Acc\\ (\%)\end{tabular}             & F1                        & \begin{tabular}[c]{@{}c@{}}Acc\\ (\%)\end{tabular}              & F1                        & \begin{tabular}[c]{@{}c@{}}Acc\\ (\%)\end{tabular}              & F1             \\ \midrule
CTransPath & KNN                 & 62.85           & 0.618          & 69.93           & 0.684          & 73.77            & 0.726           & {\textbf{92.15}}     & {\textbf{0.919}}          & 96.58                 & 0.965                     & 95.42                 & 0.954             \\
           & MatchingNet         & 24.69           & 0.111          & 49.69           & 0.433          & 44.27            & 0.388           & 32.12                & 0.232                     & 47.71                 & 0.407                     & 62.62                 & 0.597             \\
           & ProtoNet            & 25.60           & 0.231          & 60.58           & 0.586          & 65.13            & 0.644           & 51.27                & 0.483                     & 77.22                 & 0.771                     & 78.08                 & 0.778             \\
           & \texttt{Baseline}   & \textbf{64.96}  & \textbf{0.632} & \textbf{81.52}  & \textbf{0.813} & \textbf{85.25}   & \textbf{0.853}  & 91.27                & 0.909                     & 95.77                 & 0.958                     & 96.09                 & 0.961             \\
           & \texttt{Baseline}++ & 61.93           & 0.608          & 77.88           & 0.775          & 81.34            & 0.814           & 91.50                & 0.911                     & {\textbf{96.93}}      & {\textbf{0.969}}          & {\textbf{96.59}}      & {\textbf{0.966}}  \\ \midrule
Lunit      & KNN                 & 63.00           & 0.614          & 65.99           & 0.636          & 73.56            & 0.728           & {\textbf{86.06}}     & {\textbf{0.849}}          & 95.71                 & 0.956                     & 93.52                 & 0.934             \\
           & MatchingNet         & 33.56           & 0.304          & 29.26           & 0.233          & 49.51            & 0.392           & 36.01                & 0.349                     & 34.51                 & 0.296                     & 50.29                 & 0.416             \\
           & ProtoNet            & 50.27           & 0.480          & 60.25           & 0.592          & 65.62            & 0.647           & 71.42                & 0.701                     & 84.57                 & 0.844                     & 81.27                 & 0.812             \\
           & \texttt{Baseline}   & \textbf{65.19}  & \textbf{0.636} & \textbf{80.22}  & \textbf{0.800} & \textbf{83.82}   & \textbf{0.838}  & 86.05                & 0.848                     & {\textbf{96.63}}      & {\textbf{0.966}}          & {\textbf{96.08}}      & {\textbf{0.960}}  \\
           & \texttt{Baseline}++ & 64.05           & 0.624          & 79.78           & 0.794          & 82.37            & 0.823           & 85.84                & 0.846                     & 95.37                 & 0.953                     & 95.13                 & 0.951             \\ \midrule
Phikon     & KNN                 & 72.30           & 0.716          & 77.12           & 0.757          & 80.59            & 0.792           & 86.57                & 0.857                     & 94.22                 & 0.941                     & 94.92                 & 0.949             \\
           & MatchingNet         & 31.27           & 0.292          & 37.93           & 0.335          & 58.34            & 0.532           & 53.27                & 0.507                     & 59.76                 & 0.569                     & 59.29                 & 0.546             \\
           & ProtoNet            & 55.36           & 0.523          & 58.78           & 0.574          & 68.63            & 0.678           & 67.30                & 0.666                     & 78.40                 & 0.779                     & 81.10                 & 0.810             \\
           & \texttt{Baseline}   & \textbf{73.56}  & \textbf{0.727} & \textbf{84.57}  & \textbf{0.845} & 86.05            & 0.860           & 86.22                & 0.852                     & 96.84                 & 0.968                     & 97.28                 & 0.973             \\
           & \texttt{Baseline}++ & 72.33           & 0.716          & 82.84           & 0.824          & \textbf{87.21}   & \textbf{0.871}  & {\textbf{86.73}}     & {\textbf{0.859}}          & {\textbf{97.18}}      & {\textbf{0.972}}          & {\textbf{97.61}}      & {\textbf{0.976}}  \\ \midrule
UNI        & KNN                 & 63.57           & 0.618          & 72.39           & 0.718          & 73.70            & 0.719           & 87.54                & 0.866                     & 96.99                 & 0.970                     & 97.16                 & 0.971             \\
           & MatchingNet         & 42.64           & 0.410          & 43.80           & 0.350          & 46.83            & 0.383           & 61.17                & 0.578                     & 51.86                 & 0.430                     & 50.90                 & 0.440             \\
           & ProtoNet            & 54.03           & 0.513          & 62.45           & 0.615          & 62.32            & 0.619           & 69.70                & 0.694                     & 79.52                 & 0.793                     & 80.11                 & 0.798             \\
           & \texttt{Baseline}   & \textbf{64.43}  & \textbf{0.629} & 80.16           & 0.800          & 82.85            & 0.828           & 88.18                & 0.872                     & 98.08                 & 0.981                     & 98.39                 & 0.984             \\
           & \texttt{Baseline}++ & 63.74           & 0.619          & \textbf{80.35}  & \textbf{0.802} & \textbf{82.96}   & \textbf{0.829}  & {\textbf{88.36}}     & {\textbf{0.874}}          & {\textbf{98.37}}      & {\textbf{0.984}}          & {\textbf{98.76}}      & {\textbf{0.988}}  \\ \bottomrule
\end{tabular}
\end{adjustbox}
\end{table}

\begin{table}[]
\caption{Few-shot learning results on out-domain adaptation tasks.}
\label{tab:fewshot_out}
\centering
\begin{adjustbox}{max width=\textwidth}
\begin{tabular}{@{}llcccccccccccccccccc@{}}
\toprule
               &                     & \multicolumn{6}{c|}{Prostate cancer grading}                                                                                & \multicolumn{6}{c|}{Gastric cancer sub-typing}                                                                                 & \multicolumn{6}{c}{Breast cancer detection}                                                                           \\ \midrule
               &                     & \multicolumn{2}{c}{1-shot}      & \multicolumn{2}{c}{5-shot}      & \multicolumn{2}{c|}{10-shot}                         & \multicolumn{2}{c}{1-shot}      & \multicolumn{2}{c}{5-shot}      & \multicolumn{2}{c|}{10-shot}                         & \multicolumn{2}{c}{1-shot}    & \multicolumn{2}{c}{5-shot}    & \multicolumn{2}{c}{10-shot}                       \\ \midrule
    Model      & Method              & \begin{tabular}[c]{@{}c@{}}Acc\\ (\%)\end{tabular}      & F1             & \begin{tabular}[c]{@{}c@{}}Acc\\ (\%)\end{tabular}      & F1             & \begin{tabular}[c]{@{}c@{}}Acc\\ (\%)\end{tabular}      & \multicolumn{1}{c|}{F1}             & \begin{tabular}[c]{@{}c@{}}Acc\\ (\%)\end{tabular}      & F1             & \begin{tabular}[c]{@{}c@{}}Acc\\ (\%)\end{tabular}      & F1             & \begin{tabular}[c]{@{}c@{}}Acc\\ (\%)\end{tabular}      & \multicolumn{1}{c|}{F1}             & \begin{tabular}[c]{@{}c@{}}Acc\\ (\%)\end{tabular}        & F1               & \begin{tabular}[c]{@{}c@{}}Acc\\ (\%)\end{tabular}         & F1              & \begin{tabular}[c]{@{}c@{}}Acc\\ (\%)\end{tabular}         & F1                 \\ \midrule
    CTransPath & KNN                 & 44.67          & 0.421          & 65.40          & 0.637          & 68.55          & \multicolumn{1}{c|}{0.662}          & 62.81          & 0.617          & 72.22          & 0.720          & 75.45          & \multicolumn{1}{c|}{0.750}          & 55.26            & 0.545            & 62.42            & 0.624            & 71.39            & 0.697              \\
               & MatchingNet         & 23.89          & 0.121          & 33.85          & 0.274          & 34.92          & \multicolumn{1}{c|}{0.266}          & 26.20          & 0.149          & 36.18          & 0.307          & 42.89          & \multicolumn{1}{c|}{0.358}          & 25.59            & 0.128            & 34.10            & 0.259            & 37.45            & 0.289              \\
               & ProtoNet            & 27.66          & 0.254          & 32.60          & 0.298          & 32.09          & \multicolumn{1}{c|}{0.291}          & 32.17          & 0.307          & 46.97          & 0.452          & 51.43          & \multicolumn{1}{c|}{0.497}          & 35.09            & 0.328            & 38.19            & 0.368            & 46.27            & 0.436              \\
               & \texttt{Baseline}   & \textbf{46.73} & \textbf{0.439} & \textbf{74.44} & \textbf{0.737} & 72.53          & \multicolumn{1}{c|}{0.720}          & \textbf{66.68} & \textbf{0.654} & \textbf{78.93} & \textbf{0.789} & \textbf{84.82} & \multicolumn{1}{c|}{\textbf{0.848}} & {\textbf{57.40}} & {\textbf{0.569}} & {\textbf{72.78}} & {\textbf{0.723}} & {\textbf{79.15}} & {\textbf{0.788}}   \\
               & \texttt{Baseline}++ & 44.39          & 0.417          & 70.91          & 0.699          & \textbf{73.46} & \multicolumn{1}{c|}{\textbf{0.728}} & 61.73          & 0.604          & 76.49          & 0.764          & 83.49          & \multicolumn{1}{c|}{0.836}          & 54.78            & 0.541            & 72.29            & 0.718            & 77.48            & 0.769              \\ \midrule
    Lunit      & KNN                 & 45.71          & 0.429          & 63.45          & 0.614          & 65.65          & \multicolumn{1}{c|}{0.636}          & 67.00          & 0.667          & 77.82          & 0.773          & 79.73          & \multicolumn{1}{c|}{0.798}          & 59.65            & 0.585            & 62.56            & 0.617            & 71.69            & 0.701              \\
               & MatchingNet         & 22.54          & 0.219          & 25.97          & 0.200          & 26.85          & \multicolumn{1}{c|}{0.190}          & 35.35          & 0.340          & 38.08          & 0.317          & 40.75          & \multicolumn{1}{c|}{0.329}          & 38.77            & 0.370            & 44.76            & 0.406            & 40.75            & 0.335              \\
               & ProtoNet            & 29.03          & 0.262          & 33.20          & 0.319          & 37.26          & \multicolumn{1}{c|}{0.348}          & 39.12          & 0.352          & 48.44          & 0.470          & 56.46          & \multicolumn{1}{c|}{0.548}          & 36.25            & 0.346            & 42.42            & 0.405            & 43.16            & 0.416              \\
               & \texttt{Baseline}   & \textbf{48.18} & \textbf{0.461} & 72.03          & 0.712          & 70.18          & \multicolumn{1}{c|}{0.693}          & 68.50          & 0.681          & 79.69          & 0.798          & 86.03          & \multicolumn{1}{c|}{0.860}          & {\textbf{59.87}} & {\textbf{0.587}} & {\textbf{72.65}} & {\textbf{0.714}} & {\textbf{78.23}} & {\textbf{0.779}}   \\
               & \texttt{Baseline}++ & 45.95          & 0.435          & \textbf{73.20} & \textbf{0.725} & \textbf{72.14} & \multicolumn{1}{c|}{\textbf{0.710}} & \textbf{68.52} & \textbf{0.680} & \textbf{81.27} & \textbf{0.811} & \textbf{86.51} & \multicolumn{1}{c|}{\textbf{0.865}} & 58.90            & 0.578            & 72.28            & 0.711            & 77.37            & 0.770              \\ \midrule
    Phikon     & KNN                 & 45.81          & \textbf{0.435} & 64.75          & 0.616          & 67.88          & \multicolumn{1}{c|}{0.656}          & 66.27          & 0.657          & 71.95          & 0.715          & 75.85          & \multicolumn{1}{c|}{0.757}          & {\textbf{60.42}} & {\textbf{0.598}} & 65.05            & 0.644            & 72.27            & 0.717              \\
               & MatchingNet         & 24.00          & 0.223          & 26.01          & 0.201          & 26.76          & \multicolumn{1}{c|}{0.168}          & 41.35          & 0.395          & 48.09          & 0.451          & 38.78          & \multicolumn{1}{c|}{0.303}          & 40.95            & 0.399            & 43.65            & 0.382            & 34.79            & 0.293              \\
               & ProtoNet            & 29.53          & 0.268          & 33.65          & 0.325          & 35.12          & \multicolumn{1}{c|}{0.337}          & 38.46          & 0.367          & 48.83          & 0.477          & 58.81          & \multicolumn{1}{c|}{0.581}          & 41.92            & 0.388            & 42.48            & 0.402            & 51.66            & 0.487              \\
               & \texttt{Baseline}   & 45.48          & 0.419          & 73.02          & 0.725          & 73.72          & \multicolumn{1}{c|}{0.730}          & \textbf{67.71} & \textbf{0.675} & 79.20          & 0.792          & 84.52          & \multicolumn{1}{c|}{0.844}          & 59.31            & 0.588            & 73.37            & 0.730            & {\textbf{80.28}} & {\textbf{0.802}}   \\
               & \texttt{Baseline}++ & \textbf{45.94} & 0.427          & \textbf{75.28} & \textbf{0.746} & \textbf{75.30} & \multicolumn{1}{c|}{\textbf{0.745}} & 66.39          & 0.659          & \textbf{80.29} & \textbf{0.803} & \textbf{85.41} & \multicolumn{1}{c|}{\textbf{0.854}} & 59.08            & 0.585            & {\textbf{74.44}} & {\textbf{0.739}} & 78.09            & 0.780              \\ \midrule
    UNI        & KNN                 & 47.02          & 0.427          & 63.32          & 0.610          & 66.20          & \multicolumn{1}{c|}{0.631}          & \textbf{66.48} & \textbf{0.662} & 76.29          & 0.763          & 81.60          & \multicolumn{1}{c|}{0.815}          & {\textbf{63.16}} & {\textbf{0.622}} & 71.17            & 0.707            & 79.41            & 0.787              \\
               & MatchingNet         & 24.24          & 0.224          & 29.30          & 0.229          & 25.84          & \multicolumn{1}{c|}{0.193}          & 37.99          & 0.364          & 42.22          & 0.382          & 41.19          & \multicolumn{1}{c|}{0.334}          & 43.80            & 0.402            & 42.05            & 0.348            & 36.08            & 0.285              \\
               & ProtoNet            & 32.19          & 0.295          & 29.90          & 0.274          & 37.16          & \multicolumn{1}{c|}{0.356}          & 35.26          & 0.339          & 50.63          & 0.485          & 50.77          & \multicolumn{1}{c|}{0.491}          & 34.52            & 0.319            & 37.53            & 0.357            & 43.82            & 0.420              \\
               & \texttt{Baseline}   & \textbf{47.96} & \textbf{0.433} & 72.35          & 0.715          & 71.24          & \multicolumn{1}{c|}{0.699}          & 65.09          & 0.648          & \textbf{82.32} & \textbf{0.824} & \textbf{89.34} & \multicolumn{1}{c|}{\textbf{0.894}} & 62.89            & 0.619            & 76.87            & 0.760            & 82.08            & 0.820              \\
               & \texttt{Baseline}++ & 47.21          & 0.430          & \textbf{73.44} & \textbf{0.726} & \textbf{74.10} & \multicolumn{1}{c|}{\textbf{0.728}} & 66.19          & 0.659          & 82.03          & 0.821          & 89.33          & \multicolumn{1}{c|}{0.894}          & 62.92            & 0.619            & {\textbf{78.22}} & {\textbf{0.774}} & {\textbf{82.94}} & {\textbf{0.830}}   \\ \bottomrule
\end{tabular}
\end{adjustbox}
\end{table}

\begin{table}[]
\caption{Computational complexity of foundation models and few-shot learning methods with varying shot counts.}
\label{tab:model_complexity_fewshot}
\centering
\begin{adjustbox}{max width=\textwidth}
\begin{tabular}{@{}llccccccccccccc@{}}
\toprule
\multirow{3}{*}{Model} & \multirow{3}{*}{Method} & \multirow{3}{*}{\# Params (M)}  & \multicolumn{4}{c|}{1-shot}                                              & \multicolumn{4}{c|}{5-shot}                                               & \multicolumn{4}{c}{10-shot}                                       \\ \cmidrule(l){4-15}
                       &                         &                                     & \multicolumn{2}{c}{Memory (MB)}  & \multicolumn{2}{c|}{Time (sec)} & \multicolumn{2}{c}{Memory (MB)}   & \multicolumn{2}{c|}{Time (sec)}       & \multicolumn{2}{c}{Memory (MB)} & \multicolumn{2}{c}{Time (sec)}  \\ \cmidrule(l){4-15}
                       &                         &                                     & Train    & Test          & Train & \multicolumn{1}{c|}{Test}       & Train & Test          & Train     & \multicolumn{1}{c|}{Test}             & Train     & Test          & Train     & Test                      \\ \midrule
CTransPath             & KNN                     & 27.523                              & -        & 3,675         & -     & \multicolumn{1}{c|}{1.83}       & -       & 3,817       & -         & \multicolumn{1}{c|}{2.26}             & -         & 3,773         & -         & 2.86                      \\
                       & ProtoNet                & 27.523                              & 9,319    & 3,675         & 2.27  & \multicolumn{1}{c|}{1.82}       & 11,049  & 3,817       & 2.28      & \multicolumn{1}{c|}{2.26}             & 13,123    & 3,773         & 2.85      & 2.27                      \\
                       & MatchingNet             & 27.523                              & 9,319    & 3,675         & 2.66  & \multicolumn{1}{c|}{2.35}       & 11,049  & 3,817       & 2.70      & \multicolumn{1}{c|}{2.37}             & 13,123    & 3,773         & 2.68      & 2.80                      \\
                       &\texttt{Baseline}        & 27.526                              & -        & 3,683         & -     & \multicolumn{1}{c|}{204.67}     & -       & 3,781       & -         & \multicolumn{1}{c|}{205.14}           & -         & 3,825         & -         & 205.12                    \\
                       & \texttt{Baseline}++     & 27.526                              & -        & 3,683         & -     & \multicolumn{1}{c|}{218.47}     & -       & 3,781       & -         & \multicolumn{1}{c|}{222.39}           & -         & 3,825         & -         & 224.26                    \\ \midrule
Lunit                  & KNN                     & 21.667                              & -        & 2,183         & -     & \multicolumn{1}{c|}{1.82}       & -       & 2,243       & -         & \multicolumn{1}{c|}{1.82}             & -         & 2,373         & -         & 1.83                      \\
                       & ProtoNet                & 21.667                              & 6,641    & 2,183         & 2.28  & \multicolumn{1}{c|}{1.83}       & 7,777   & 2,243       & 2.71      & \multicolumn{1}{c|}{1.86}             & 9,293     & 2,373         & 2.69      & 2.70                      \\
                       & MatchingNet             & 21.667                              & 6,641    & 2,183         & 1.83  & \multicolumn{1}{c|}{1.87}       & 7,759   & 2,243       & 2.27      & \multicolumn{1}{c|}{1.92}             & 9,297     & 2,373         & 2.66      & 2.37                      \\
                       &\texttt{Baseline}        & 21.669                              & -        & 2,191         & -     & \multicolumn{1}{c|}{203.42}     & -       & 2,251       & -         & \multicolumn{1}{c|}{203.35}           & -         & 2,381         & -         & 205.14                    \\
                       & \texttt{Baseline}++     & 21.669                              & -        & 2,191         & -     & \multicolumn{1}{c|}{218.69}     & -       & 2,251       & -         & \multicolumn{1}{c|}{219.45}           & -         & 2,381         & -         & 221.05                    \\ \midrule
Phikon                 & KNN                     & 85.802                              & -        & 2,871         & -     & \multicolumn{1}{c|}{1.83}       & -       & 2,951       & -         & \multicolumn{1}{c|}{2.26}             & -         & 3,103         & -         & 2.27                      \\
                       & ProtoNet                & 85.802                              & 12,263   & 2,871         & 2.28  & \multicolumn{1}{c|}{1.86}       & 14,493  & 2,951       & 2.30      & \multicolumn{1}{c|}{2.30}             & 17,893    & 3,103         & 2.71      & 2.25                      \\
                       & MatchingNet             & 85.802                              & 12,263   & 2,871         & 1.86  & \multicolumn{1}{c|}{2.38}       & 14,493  & 2,951       & 2.30      & \multicolumn{1}{c|}{2.40}             & 17,893    & 3,103         & 2.84      & 2.83                      \\
                       &\texttt{Baseline}        & 85.805                              & -        & 2,879         & -     & \multicolumn{1}{c|}{203.53}     & -       & 2,959       & -         & \multicolumn{1}{c|}{208.41}           & -         & 3,111         & -         & 209.38                    \\
                       & \texttt{Baseline}++     & 85.805                              & -        & 2,879         & -     & \multicolumn{1}{c|}{203.20}     & -       & 2,959       & -         & \multicolumn{1}{c|}{203.32}           & -         & 3,111         & -         & 206.58                    \\ \midrule
UNI                    & KNN                     & 303.355                             & -        & 3,677         & -     & \multicolumn{1}{c|}{1.84}       & -       & 3,683       & -         & \multicolumn{1}{c|}{1.85}             & -         & 3,897         & -         & 1.85                      \\
                       & ProtoNet                & 303.355                             & 28,680   & 3,677         & 3.63  & \multicolumn{1}{c|}{1.84}       & 34,161  & 3,683       & 4.11      & \multicolumn{1}{c|}{1.87}             & 42,000    & 3,897         & 4.16      & 2.31                      \\
                       & MatchingNet             & 303.355                             & 28,680   & 3,677         & 2.61  & \multicolumn{1}{c|}{1.94}       & 34,161  & 3,683       & 2.72      & \multicolumn{1}{c|}{2.39}             & 42,000    & 3,897         & 4.13      & 2.39                      \\
                       &\texttt{Baseline}        & 303.359                             & -        & 3,685         & -     & \multicolumn{1}{c|}{203.35}     & -       & 3,691       & -         & \multicolumn{1}{c|}{203.53}           & -         & 3,905         & -         & 205.81                    \\
                       & \texttt{Baseline}++     & 303.359                             & -        & 3,685         & -     & \multicolumn{1}{c|}{202.77}     & -       & 3,691       & -         & \multicolumn{1}{c|}{204.09}           & -         & 3,905         & -         & 203.61                    \\ \bottomrule
\end{tabular}
\end{adjustbox}
\end{table}

\section{Discussion}
This study provides a comprehensive evaluation of pathology-specific foundation models, assessing their adaptability across various datasets and tasks through consistency and flexibility assessment scenarios. The experimental results offer key insights into how these foundation models can be effectively fine-tuned and applied to various pathology tasks under diverse conditions.

Our analyses revealed that UNI exhibited the most stable performance among the pathology-specific four models in the consistency assessment scenario. The success of UNI is primarily attributable to its larger model size and extensive training data. UNI employed the ViT-L architecture as its backbone and was pretrained using DINOv2 on the MASS-100k dataset. With 303.355 million parameters, UNI has the largest number of parameters among the four foundation models and was trained on the most extensive dataset. This combination of vast data and numerous parameters played a critical role in obtaining superior generalization performance compared to other foundation models. 
However, the large model size introduces a trade-off between performance and complexity. Due to the extensive parameter count, UNI consumes more computational resources during training compared to other foundation models. Given the continuous increase in dataset sizes across numerous tasks in pathology, the high complexity of UNI can be a practical concern. 
This issue can be somewhat mitigated through the use of PEFT. PEFT showed superior performance in the consistency assessment scenario across various datasets and tasks compared to other adaptation strategies. Strength of PEFT lies in its ability to achieve excellent performance while minimizing computational resources, as opposed to traditional fine-tuning methods such as linear probing or full fine-tuning. This makes PEFT a highly efficient and effective approach for both development and deployment in real-world clinical environments.

In the flexibility assessment scenario, FSL methods such as \texttt{Baseline} and \texttt{Baseline}++ proved effective in adapting pathology-specific foundation models to new tasks with limited data. These methods preserved capabilities of pre-trained models while efficiently adapting to new, unseen data, requiring only minimal modifications to the original models. 
On the other hand, approaches that involve fine-tuning the entire model, like MatchingNet and ProtoNet, performed worse than KNN. 
As training data is extremely limited (i.e., 1-shot out-domain scenarios), KNN sometimes outperformed \texttt{Baseline} and \texttt{Baseline}++, indicating that the prior knowledge of the foundation models is sufficient to handle new tasks. 
However, as the number of shots increases, the adoption of \texttt{Baseline} and \texttt{Baseline}++ provided superior performance. 
This suggests that while the foundation models can handle new tasks with minimal data, they still benefit from additional data and specialized FSL methods to further enhance their performance.

Comparing the computational complexity of the four pathology-specific foundation models, the test time per image (equivalent to the test time with a batch size of 1) was similar across foundation models, whereas the test time per batch varied. 
This difference is attributable to the computational efficiency in handling larger quantities of data in a batch. With smaller batch sizes, all intermediate values generated during model operations can be stored in on-chip memory, reducing the need for off-chip memory access and thus minimizing any significant increase in processing time across foundation models. 
However, as larger batch sizes grow and models become larger, the test time surges significantly, likely due to factors such as memory bandwidth and hierarchy, the size of intermediate values, and data access patterns. 
For example, the test time (and training time) per batch for UNI was more than nine times higher than that for Lunit. 
The larger model size of UNI needs processing a greater number of parameters and intermediate values, requiring more frequent off-chip memory accesses, which leads to increased processing time. Conversely, Lunit demonstrated a relatively smaller increase in execution time per batch, owing to its smaller number of parameters and lower memory usage. 
Thus, more complex models experience a sharp rise in processing time with larger batch sizes, whereas the difference in execution time between models remains minimal with smaller batch sizes. These observations are consistent with the previous findings described in \citep{park2018deep}. 

Our study emphasizes the importance of selecting appropriate adaptation strategies for pathology-specific foundation models. 
Fine-tuning while preserving existing knowledge is crucial for effectively successfully applying the foundation models to a variety of tasks. 
Though some studies highlighted the zero-shot capabilities of foundation models, particularly visual-language foundation models \citep{huang2023visual,lu2024visual}, the results in the flexibility assessment scenarios suggest that utilizing additional small data is still beneficial. FSL methods, however, have not yet been fully studied and developed for pathology-specific foundation models, presenting an area for further exploration. Moreover, the size of foundation models tends to increase, so do the execution time and memory requirements for both training and testing, which could pose a significant barrier to their practical use in clinical settings, in regard to the enormous number of tasks in pathology. 

There are limitations in our study. 
First, our study is limited to patch-level pathology image classification, which does not encompass various aspects of pathology image analysis. It is necessary to expand our study to other tasks such as whole slide-level classification \citep{kim2023paip}, cell segmentation \citep{doan2022sonnet}, and etc. 
Second, five fine-tuning and five FSL methods are employed in this study. While PEFT and \texttt{Baseline}/\texttt{Baseline}++ consistently showed advantages in this benchmark, performance varied depending on the specific foundation model and dataset/task. For instance, although PEFT generally yielded superior results, partial fine-tuning still showed competitive performance in certain tasks, such as prostate cancer grading. Optimizing a foundation model for a particular dataset/task requires a further investigation. This issue is beyond the scope of this study and will be explored in future research.
Third, in the flexibility assessment scenario, colorectal tissue datasets are used for $X_{Meta-train}$ and $X_{Meta-val}$, while $X_{Meta-test}$ includes colon, lung, prostate, gastric, and breast tissues. The relationship between colorectal tissues with other tissue types may vary, which could influence the performance in adaptation tasks. We leave further exploration of this issue for future study.
Last, while the consistency assessment scenario adopts multiple dataset per adaptation task, the flexibility assessment scenario involves a single dataset for each specific adaptation task. To further validate our findings on the flexibility assessment of the foundation models, an extended validation study, including multiple external datasets, needs to be followed.

\section{Conclusion}
In this study, we benchmarked four pathology-specific foundation models, focusing on adaptation strategies and performance across various scenarios. 
Fine-tuning foundation models, particularly using PEFT, proved effective in addressing data variabilities due to differences in data source and acquisition settings within the same downstream tasks. PEFT consistently outperformed other fine-tuning methods by maintaining superior performance across diverse datasets/tasks while demonstrating comparable computational complexity.
Moreover, the strength and usefulness of foundation models were validated in data-limited environments. Foundation models with KNN demonstrated the ability to adapt to unseen tasks, leveraging the prior knowledge of foundation models in pathology. The integration of specialized FSL methods, such as \texttt{Baseline} and \texttt{Baseline}++, could further enhance the generalization ability of foundation models, enabling more robust performance across various adaptation tasks.
Future study will focus on developing more advanced fine-tuning and FSL methods tailored to pathology-specific foundation models and optimizing them to further improve both performance and computational complexity.

\vspace{12pt}

\section*{Acknowledgment}
This work was supported by the National Research Foundation of Korea (NRF) (No. 2021R1A2C2014557 and No. RS-2024-00397293) and by the Ministry of Trade, Industry and Energy (MOTIE) and Korea Institute for Advancement of Technology (KIAT) through the International Cooperative R\&D program (No. P0022543).
\vspace{12pt}

\bibliography{ref}
\bibliographystyle{unsrt}
\end{document}